\documentclass[12pt, final]{l4dc2020} 

\usepackage[utf8]{inputenc} 
\usepackage[T1]{fontenc}    
\usepackage{hyperref}       
\usepackage{url}            
\usepackage{booktabs}       
\usepackage{amsfonts}       
\usepackage{nicefrac}       
\usepackage{microtype}      

\usepackage{color}
\usepackage{todonotes}

\usepackage{xspace}
\newcommand{\eg}{{\it e.g.}\xspace}

\newcommand{\sect}[1]{Sec.~\ref{#1}}
\newcommand{\fig}[1]{Fig.~\ref{#1}}
\newcommand{\eq}[1]{Eq.~\eqref{#1}}

\newcommand{\tab}[1]{Tab.~\ref{#1}}

\usepackage{wrapfig}

\usepackage{dsfont}
\usepackage{amsmath}

\renewcommand{\vec}[1]{\boldsymbol{#1}}				

\newcommand{\R}[0]{\mathds{R}}					

\newcommand{\E}{\mathds{E}}	 				

\newcommand{\prob}{{p}} 					

\newcommand{\D}[0]{\mathcal{D}} 				

\DeclareMathOperator*{\argmax}{arg\,max}

\newcommand{\traj}[0]{\vec{\tau}}

\newcommand{\cblock}[3]{
  \hspace{-1.5mm}
  \begin{tikzpicture}
    [
    node/.style={square, minimum size=10mm, thick, line width=0pt},
    ]
    \node[fill={rgb,255:red,#1;green,#2;blue,#3}] () [] {};
  \end{tikzpicture}%
}

\title[Objective Mismatch in Model-based Reinforcement Learning]{Objective Mismatch in Model-based Reinforcement Learning}
\usepackage{times}

\author{%
  \Name{Nathan Lambert} \Email{nol@berkeley.edu}  \\
  \addr University of California, Berkeley
   \AND
    \Name{Brandon Amos} \Email{bda@fb.com}\\
     \Name{Omry Yadan} \Email{omry@fb.com}\\
     \Name{Roberto Calandra} \Email{rcalandra@fb.com}\\
     \addr Facebook AI Research
}

\makeatletter
 \let\Ginclude@graphics\@org@Ginclude@graphics 
\makeatother

\begin{document}

\maketitle


\begin{abstract}
	Model-based reinforcement learning (MBRL) is a powerful framework for data-efficiently learning control of continuous tasks.
Recent work in MBRL has mostly focused on using more advanced function approximators and planning schemes, with little development of the general framework.
In this paper, we identify a fundamental issue of the standard MBRL framework -- what we call \textit{objective mismatch}.
Objective mismatch arises when one objective is optimized in the hope that a second, often uncorrelated, metric will also be optimized.
In the context of MBRL, we characterize the objective mismatch between training the forward dynamics model w.r.t.~the likelihood of the one-step ahead prediction, and the overall goal of improving performance on a downstream control task.
For example, this issue can emerge with the realization that dynamics models effective for a specific task do not necessarily need to be globally accurate, and vice versa globally accurate models might not be sufficiently accurate locally to obtain good control performance on a specific task.
In our experiments, we study this objective mismatch issue and demonstrate that the likelihood of one-step ahead predictions is not always correlated with control performance.
This observation highlights a critical limitation in the MBRL framework which will require further research to be fully understood and addressed.
We propose an initial method to mitigate the mismatch issue by re-weighting dynamics model training.
Building on it, we conclude with a discussion about other potential directions of research for addressing this issue.

\end{abstract}


\section{Introduction}
Model-based reinforcement learning (MBRL) is a popular approach for learning to control nonlinear systems that cannot be expressed analytically~\citep{bertsekas1995dynamic,sutton2018reinforcement, deisenroth2011pilco, williams2017information}.
MBRL techniques achieve the state of the art performance for continuous-control problems with access to a limited number of trials~\citep{chua2018deep,wang2019exploring} and in controlling systems given only visual observations with no observations of the original system's state~\citep{hafner2018learning,zhang2018solar}.
MBRL approaches typically learn a \textit{forward dynamics model} that predicts how the dynamical system will evolve when a set of control signals are applied.
This model is classically fit with respect to the maximum likelihood of a set of trajectories collected on the real system, and then used as part of a control algorithm to be executed on the system (e.g., model-predictive control).

In this paper, we highlight a fundamental problem in the MBRL learning scheme: the \textit{objective mismatch} issue. 
The learning of the forward dynamics model is decoupled from the subsequent controller through the optimization of two different objective functions -- prediction accuracy or loss of the single- or multi-step look-ahead prediction for the dynamics model, and task performance for the policy optimization.
While the use of log-likelihood (LL) for system identification is an historically accepted objective, it results in optimizing an objective that does not necessarily correlate to controller performance.
The contributions of this paper are to:
1) identify and formalize the problem of objective mismatch in MBRL;
2) examine the signs of and the effects of objective mismatch on simulated control tasks;
3) propose an initial mechanism to mitigate objective mismatch;
4) discuss the impact of objective mismatch and outline future directions to address this issue.


\section{Model-based Reinforcement Learning}
\label{sec:mbrl}

\begin{figure}[t]
  \centering
  \includegraphics[width=.85\linewidth]{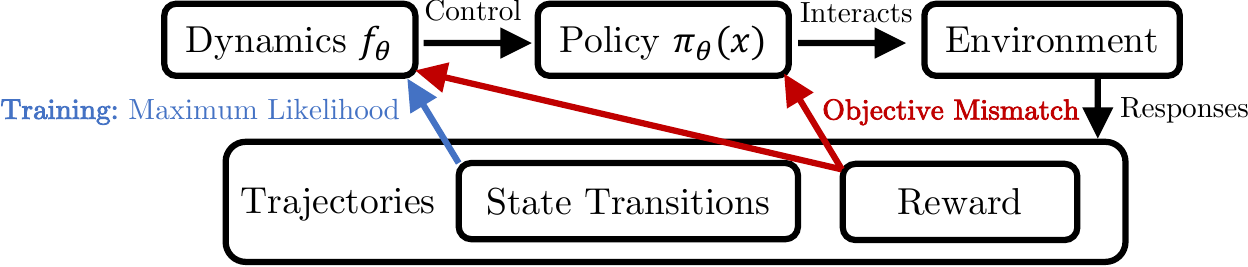}
  \caption{Objective mismatch in MBRL arises when a model is trained
    to maximize the likelihood but then used for control
    to maximize a reward signal not considered during training.
    \vspace{-15pt}
  }
  \label{fig:teaser}
\end{figure}

We now outline the MBRL formulation used in the paper.
At time $t$, we denote the state $s_t \in \R^{d_s}$, the actions $a_t \in \R^{d_a}$, and the reward $r(s_t,a_t)$. 
We say that the MBRL agent acts in an environment 
governed by a state transition distribution
$\prob(s_{t+1}|s_t,a_t)$.
We denote a parametric model~$f_\theta$ to approximate this distribution with
$\prob_\theta(s_{t+1}|s_t,a_t)$.
MBRL follows the approach of an agent acting in its environment, learning a model of said environment, and then leveraging the model to act. 
While iterating over parametric control policies, the agent collects measurements of state, action, next-state and
forms a dataset $\D=\{(s_n, a_n, s_{n}')\}_{n=1}^N$. 
With the dynamics data~$\D$, the agent learns the environment in the form of a neural network forward dynamics model, learning an approximate dynamics~$f_\theta$. 
This dynamics model is leveraged by a controller that takes in
the current state $s_t$ and returns an action sequence $a_{t:t+T}$
maximizing the expected reward $\E_{\pi_\theta(s_t)} \sum_{i=t}^{t+T}
r(s_i,a_i)$, where $T$ is the predictive horizon and
$\pi_\theta(s_t)$ is the set of state transitions induced
by the model~$p_\theta$.

In our paper, we primarily use as probabilistic neural networks designed to maximize the LL of the predicted parametric distribution $p_\theta$, denoted as $P$, or ensembles of probabilistic networks denoted $PE$, and compare to deterministic networks minimizing the mean squared error~(MSE), denoted $D$ or $DE$.
Unless otherwise stated we use the models as
in PETS~\citep{chua2018deep} with an
expectation-based trajectory planner and
a cross-entropy-method~(CEM) optimizer.


\section{Objective Mismatch and its Consequences}
\label{sec:approach}

\paragraph{The Origin of Objective Mismatch: The Subtle Differences between MBRL and System Identification}
\label{sec:origin}

Many ideas and concepts in model-based RL are rooted in the field of optimal control and system identification~\citep{Sutton1991Dyna, bertsekas1995dynamic,zhou1996robust,kirk2012optimal,bryson2018applied}.
In system identification (SI), the main idea is to use a two-step process where we first generate (optimal) elicitation trajectories~$\traj$ to fit a dynamics model (typically analytical), and subsequently we apply this model to a specific task. This particular scheme has several assumptions: 
\begin{wrapfigure}{r}{.53\textwidth}
    \centering
    \includegraphics[width=0.49\linewidth]{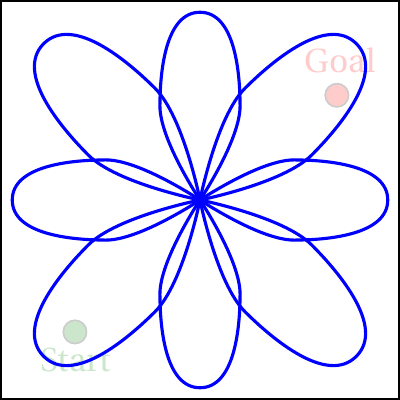}
    \hfill
    \includegraphics[width=0.49\linewidth]{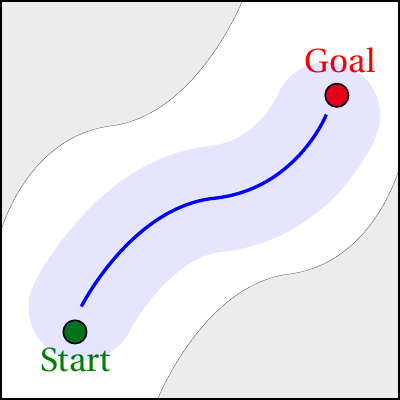}
    \vspace{-15pt}
    \caption{
    Sketches of state-action spaces. (\textit{Left}) In system identification, the elicitation trajectories are designed off-line to cover the entire state-action space. 
    (\textit{Right}) In MBRL instead, the data collected during learning is often concentrated in trajectories towards the goal, with other parts of the state-action space being largely unexplored (grey area).}
    \label{fig:space}
    \vspace{-10pt}
\end{wrapfigure}
1)~the elicitation trajectories collected cover the entire state-action space;
2)~the presence of virtually infinite amount of data; 
3)~the global and generalizable nature of the model resulting from the SI process.
With these assumptions, the theme of system identification is effectively to collect a large amount of data covering the whole state-space to create a sufficiently accurate, global model that we can deploy on any desired task, and still obtain good performance.
If these assumptions are true, using the closed-loop of MBRL should further improve performance over traditional open-loop SI~\citep{Hjalmarsson1996model}.

When adopting the idea of learning the dynamics model used in optimal control for MBRL, it is important to consider if these assumptions still hold.
The assumption of virtually infinite data is visibly in tension with the explicit goal of MBRL which is to reduce the number of interactions with the environment by being ``smart'' about the sampling of new trajectories. 
In fact, in MBRL the offline data collection performed via elicitation trajectories is largely replaced by on-policy sampling to explicitly reduce the need to collect large amount of data \citep{chua2018deep}.  
Moreover, in the MBRL setting the data will not usually cover the entire state-action space, since they are generated by optimizing one task. 
In conjunction with the use of non-parametric models, this results in learned models that are strongly biased towards capturing the distribution of the locally accurate, task-specific data.
Nonetheless, this is not an immediate issue since the MBRL setting rarely tests for generalization capabilities of the learned dynamics.
In practice, we can now see how the assumptions and goals of system identification are in contrast with the ones of MBRL.
Understanding these differences and the downstream effects on algorithmic approach is crucial to design new families of MBRL algorithms.

\paragraph{Objective Mismatch}

During the MBRL process of iteratively learning a controller, the reward signal from the environment is diluted by the training of a forward dynamics model with a independent metric, as shown in \fig{fig:teaser}.
In our experiments, we highlight that the minimization of some network training cost does not hold a strong correlation to maximization of episode reward.
As dynamic environments becoming increasingly complex in dimensionality, the assumptions of collected data distributions become weaker and over-fitting to different data poses an increased risk.

Formally, the problem of objective mismatch appears as two de-coupled optimization problems repeated over many cycles of learning, 
shown in \eq{eq:mismatch}, which could be at the cost of minimizing the final reward.
This loop becomes increasingly difficult to analyze as the dataset used for model training changes with each experimental trial -- a step that is needed to include new data from previously unexplored states.
In this paper we characterize the problems introduced by the interaction of these two optimization problems, but, for simplicity, we do not consider the interactions added by the changes in the dynamics-data distribution during the learning process. 
In addition, we discuss potential solutions, but do not make claims about the best way to do so, which is left for future work.
\begin{subequations}\label{eq:2}
\begin{gather}
    \textbf{Training: }\argmax_\theta \sum_{i=1}^N \log\prob_\theta(s_i' | s_i,a_i),   \quad \textbf{Control: }\argmax_{a_{t:t+T}} \E_{\pi_\theta(s_t)} \sum_{i=t}^{t+T} r(s_i,a_i)     \tag{\theequation a,b} 
\label{eq:mismatch}
\end{gather}
\end{subequations}

\section{Identifying Objective Mismatch}
\label{sec:experiments}

We now experimentally study the issue of objective mismatch to answer the following:
1)~Does the distribution of models obtained from running a MBRL algorithm show a strong correlation between LL and reward?
2)~Are there signs of sub-optimality in the dynamics models training process that could be limiting performance? 
3)~What model differences are reflected in reward but not in LL?  

\paragraph{Experimental Setting} 
In our experiments, we use two popular RL benchmark tasks: the cartpole (CP) and half cheetah (HC). 
For more details on these tasks, model parameters, and control properties see \citet{chua2018deep}.
We use a set of 3 different datasets to evaluate how assumptions in MBRL affect performance.
We start with high-reward, expert datasets (cartpole $r>179$, half cheetah $r>10000$) to test if on-policy performance is linked to a minimal, optimal exploration. 
The two other baselines are datasets collected on-policy with the PETS algorithm and datasets of sampled tuples representative of the entire state space. 
The experiments validate over a) many re-trained models and b) many random seeds, to account for multiple sources of stochasticity in MBRL. 
Additional details and experiments can be found at: \href{https://sites.google.com/view/mbrl-mismatch}{https://sites.google.com/view/mbrl-mismatch}.
\begin{figure}[t]
  \floatconts
    {fig:nllVr}
      {
       \vspace{-10pt}
      \caption{The distribution of dynamics models ($M_{models}= 1000,2400$ for cartpole, half cheetah) from our experiments plotting in the LL-Reward space on three datasets, with correlation coefficients $\rho$. 
      Each reward point is the mean over 10 trials.
      There is a trend of high reward to increased LL that breaks down as the datasets contain more of the state-space than only expert trajectories. 
      \vspace{-5pt}
      }
      }
      {
        \subfigure[CP Expert ($\rho=0.59$)]{\label{fig:cp-nllVr1}%
          \includegraphics[width=0.32\linewidth]{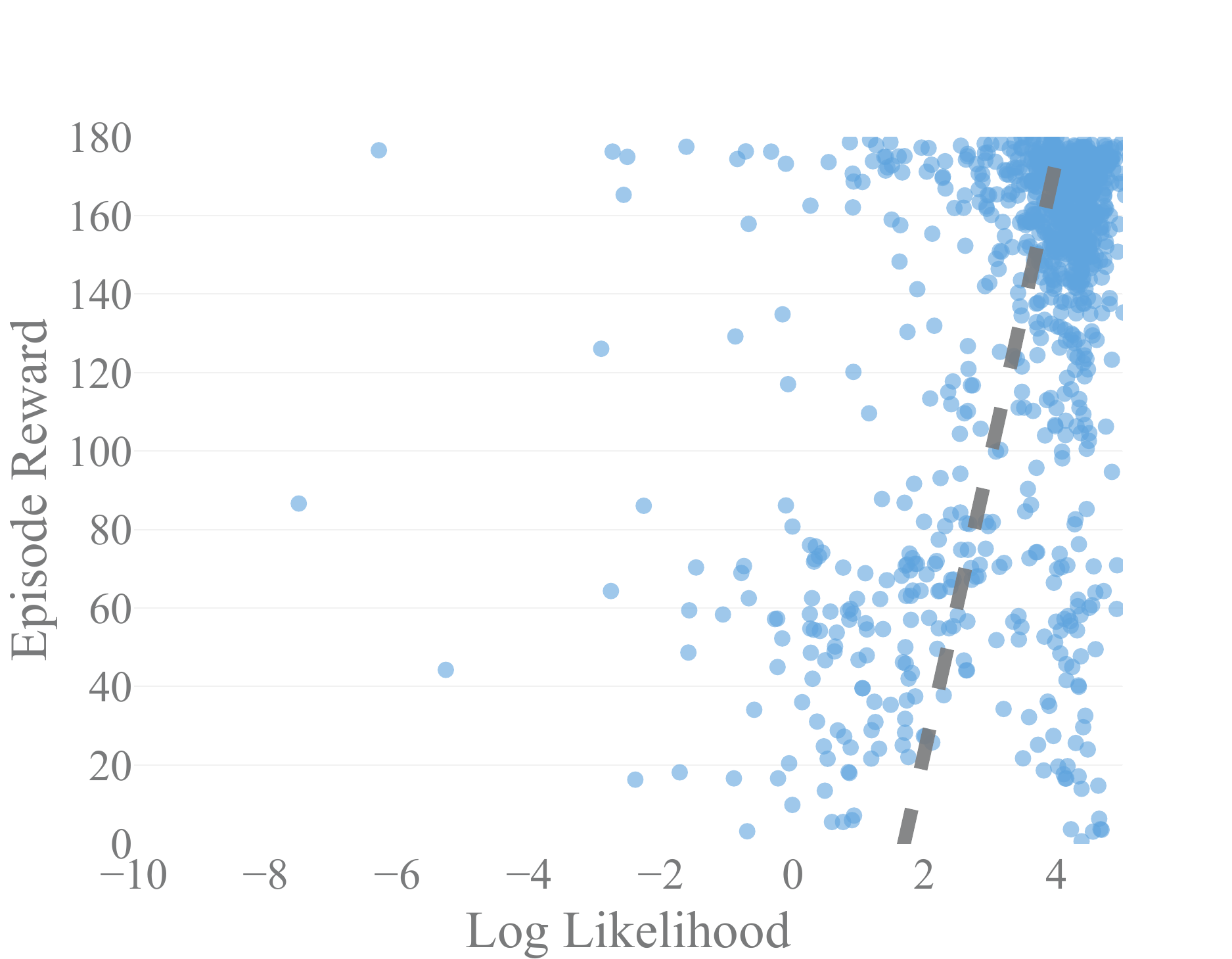}}%
        ~
        \subfigure[CP On-Policy ($\rho=0.34$)]{\label{fig:cp-nllVr2}%
          \includegraphics[width=0.32\linewidth]{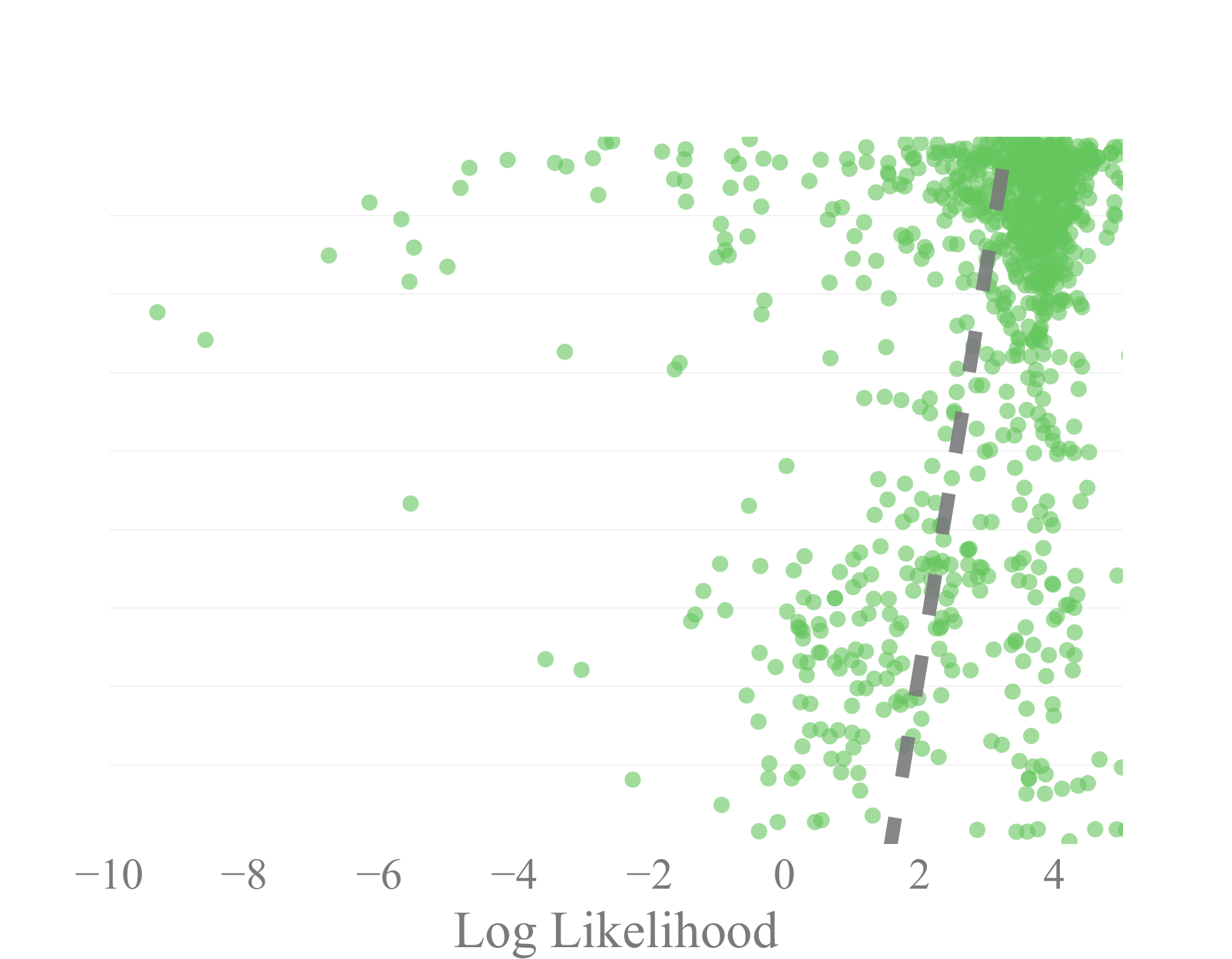}}
          ~
        \subfigure[CP Grid ($\rho=-0.06$)]{\label{fig:cp-nllVr3}%
          \includegraphics[width=0.32\linewidth]{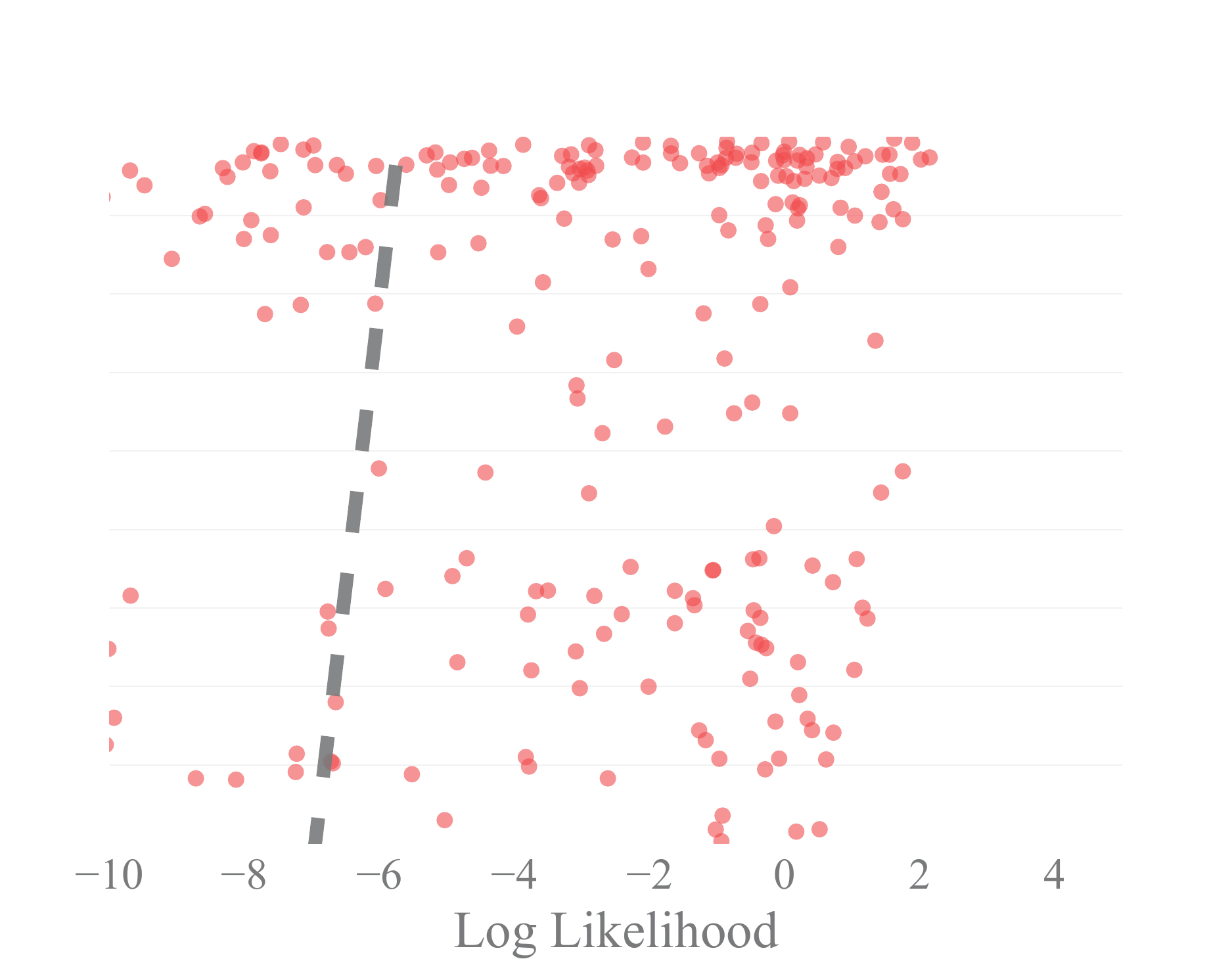}}
         \\
         \subfigure[HC Expert ($\rho=0.07$)]{\label{fig:hc-nllVr1}%
          \includegraphics[width=0.32\linewidth]{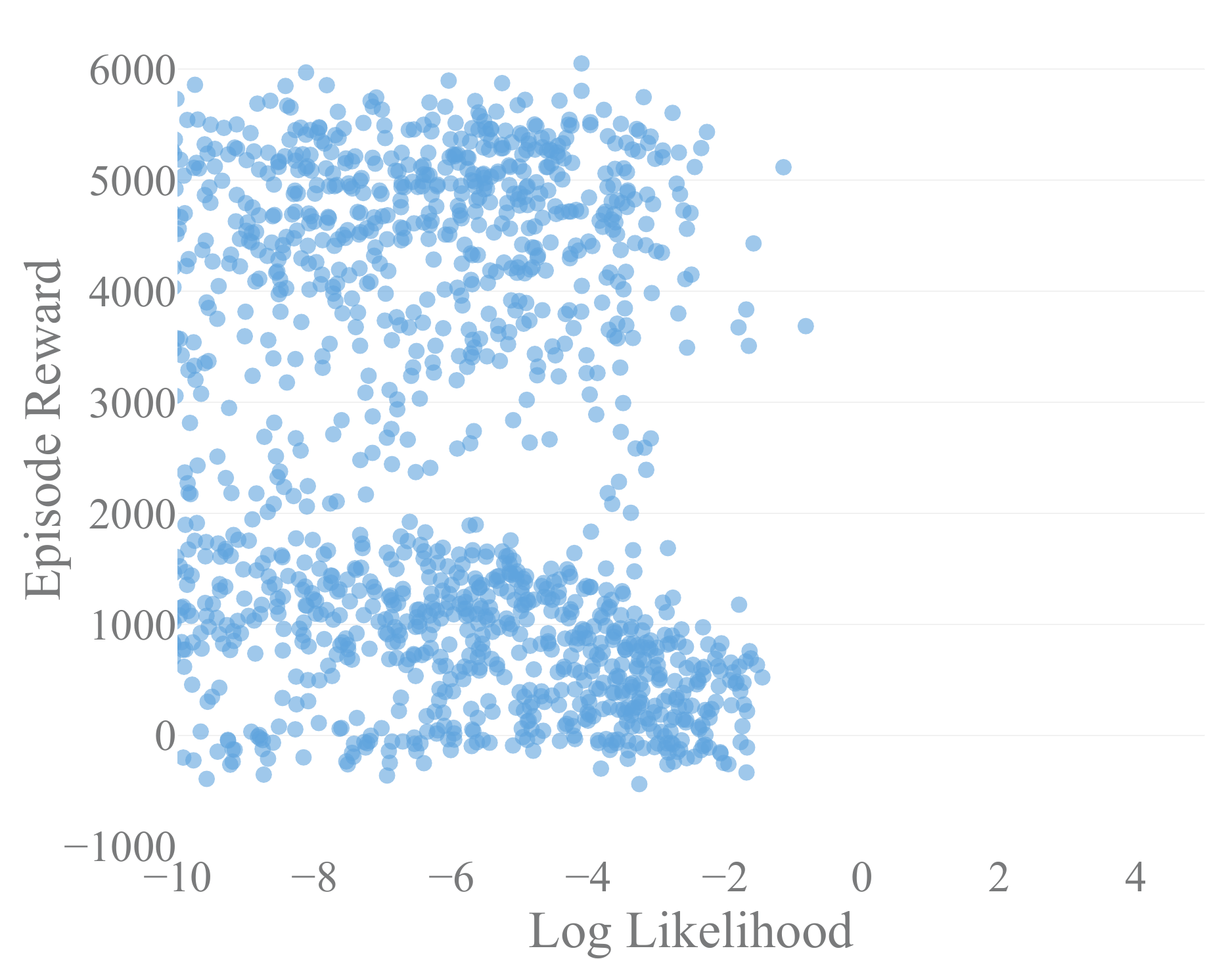}}%
        ~
        \subfigure[HC On-Policy ($\rho=0.46$)]{\label{fig:hc-nllVr2}%
          \includegraphics[width=0.32\linewidth]{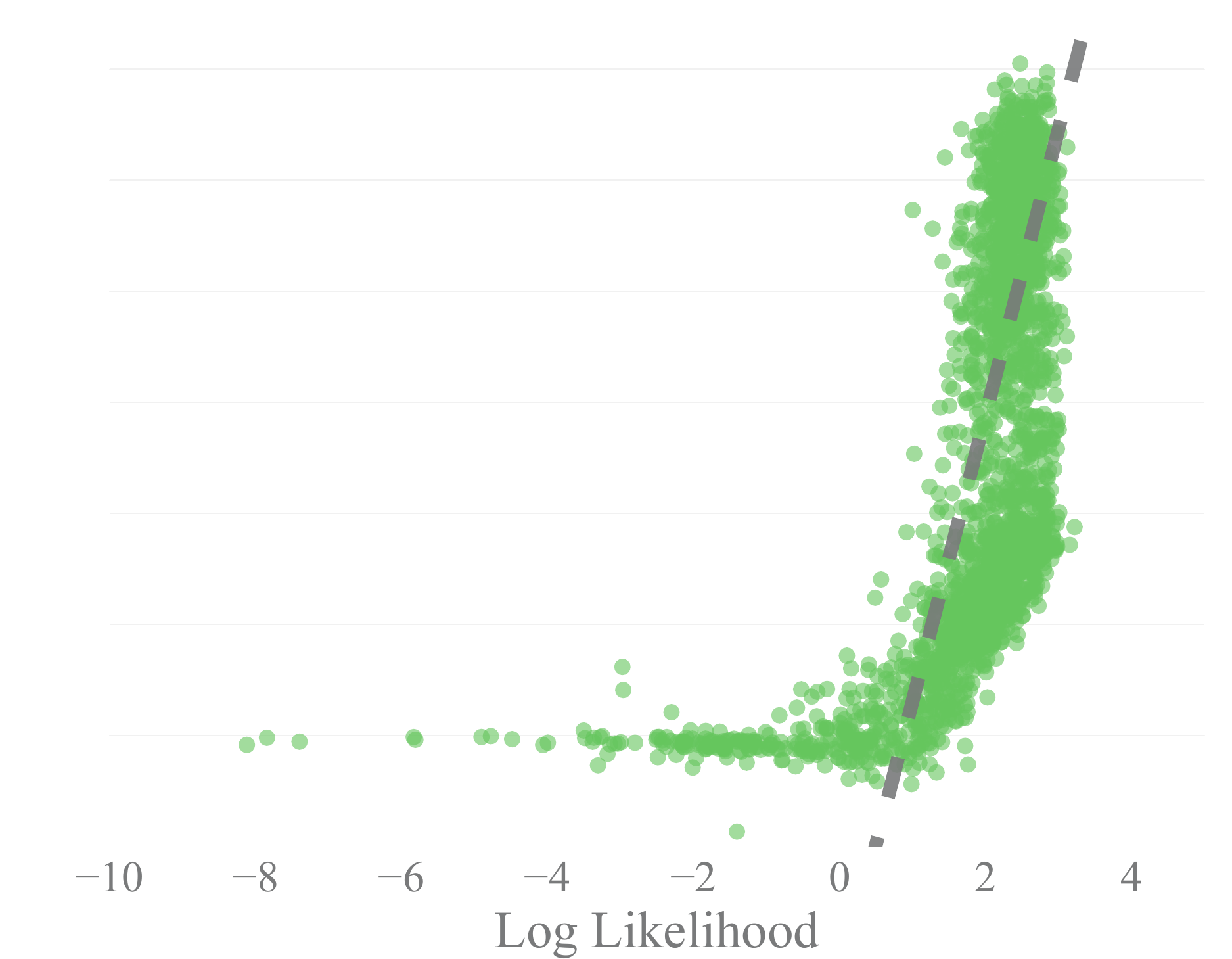}}
          ~
        \subfigure[HC Sampled  ($\rho=0.19$)]{\label{fig:hc-nllVr3}%
          \includegraphics[width=0.32\linewidth]{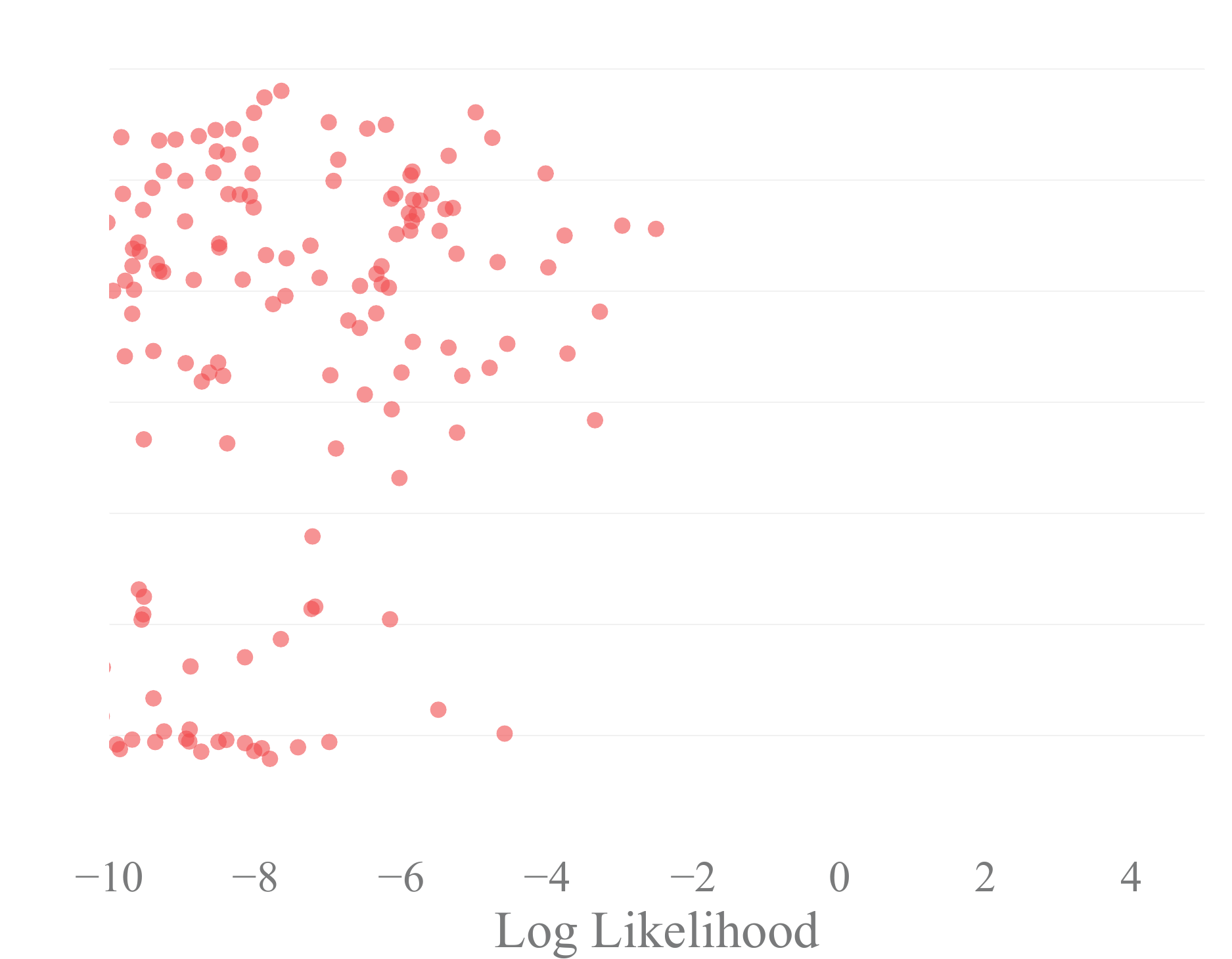}}
      }
      
\end{figure}

\subsection{Exploration of Model Loss vs Episode Reward Space}
\label{sec:nllvr}
The MBRL framework assumes a clear correlation between model accuracy and policy performance, which we challenge even in simple domains. 
We aggregated $M_{cp}=1000$ cartpole models and $M_{hc}=2400$ half cheetah models trained with PETS.
The relationships between model accuracy and reward on data representing the full state-space (grid or sampled) show no clear trend in \fig{fig:nllVr}c,f.
The distribution of rewards versus LL shown in \fig{fig:nllVr}a-c shows substantial variance and points of disagreement overshadowing a visual correlation of increased reward and LL.
This bi-model distribution on the half cheetah expert dataset, shown in  \fig{fig:hc-nllVr1}, relates to a unrecoverable failure mode in early half cheetah trials.
The contrast between \fig{fig:hc-nllVr2} and \fig{fig:nllVr}d,f shows a considerable per-dataset variation in the state-action transitions.
The grid and sampled datasets, \fig{fig:nllVr}c,f, suffer from decreased likelihood because they do not overlap greatly with on-policy data from PETS.

If the assumptions behind MBRL were fully valid, the plots should show a perfect correlation between LL and reward.
Instead these results confirm that there exists an objective mismatch which manifests as a decreased correlation between validation loss and episode reward. 
Hence, there is no guarantee that increasing the model accuracy (i.e., the LL) will also improve the control performance.

\subsection{Model Loss vs Episode Reward During Training}
\label{sec:traincollect}

\begin{figure}[t]
  \floatconts{fig:cp-tc}
  {\vspace{-10pt}
  \caption{The reward when re-evaluating the controller at each dynamics model training epoch for different dynamics models, $M=50$ per model type. 
    Even for the simple cartpole environment, 
     $D$, $DE$ fail to achieve full performance, while $P$, $PE$ reach higher performance but eventually over-fit to available data.
     The validation loss is still improving slowly at 500 epochs, not yet over-fitting.
     \vspace{-5pt}
    }
    }
    {\centering 
  \small{
    \cblock{31}{119}{180} Validation Error (\textcolor[rgb]{.12,.46,.71}{$	\bigtriangleup$}$:DE$, $PE$, \textcolor[rgb]{.12,.46,.71}{$	\Diamond$}$:D$, $P$) \quad
    \cblock{255}{69}{46} Episode Reward (\textcolor{red}{$\mathbb{X}$}$:DE, PE$, \textcolor{red}{\large{$\star$}}$:D, P$) 
    } %
        \subfigure[$P, PE$ models.]{ 
          \includegraphics[width=0.48\linewidth]{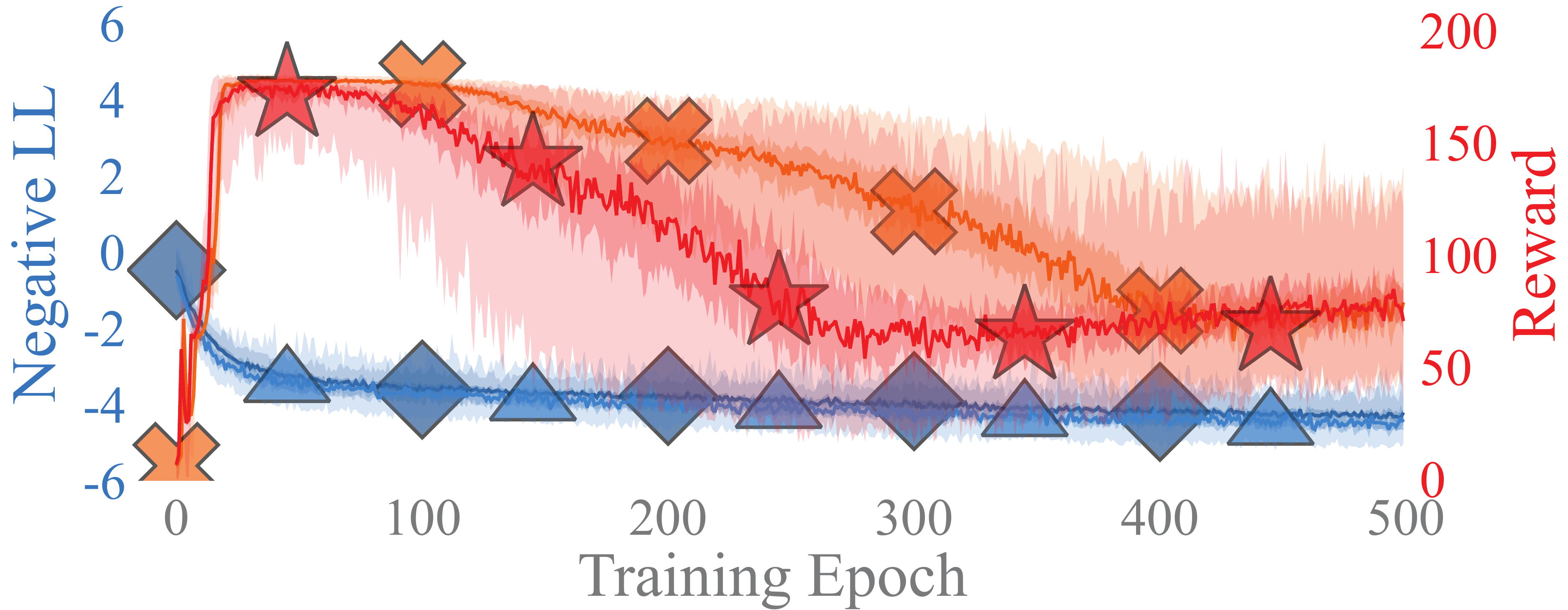}}%
        \hfill
        \subfigure[$D, DE$ models.]{ 
          \includegraphics[width=0.48\linewidth]{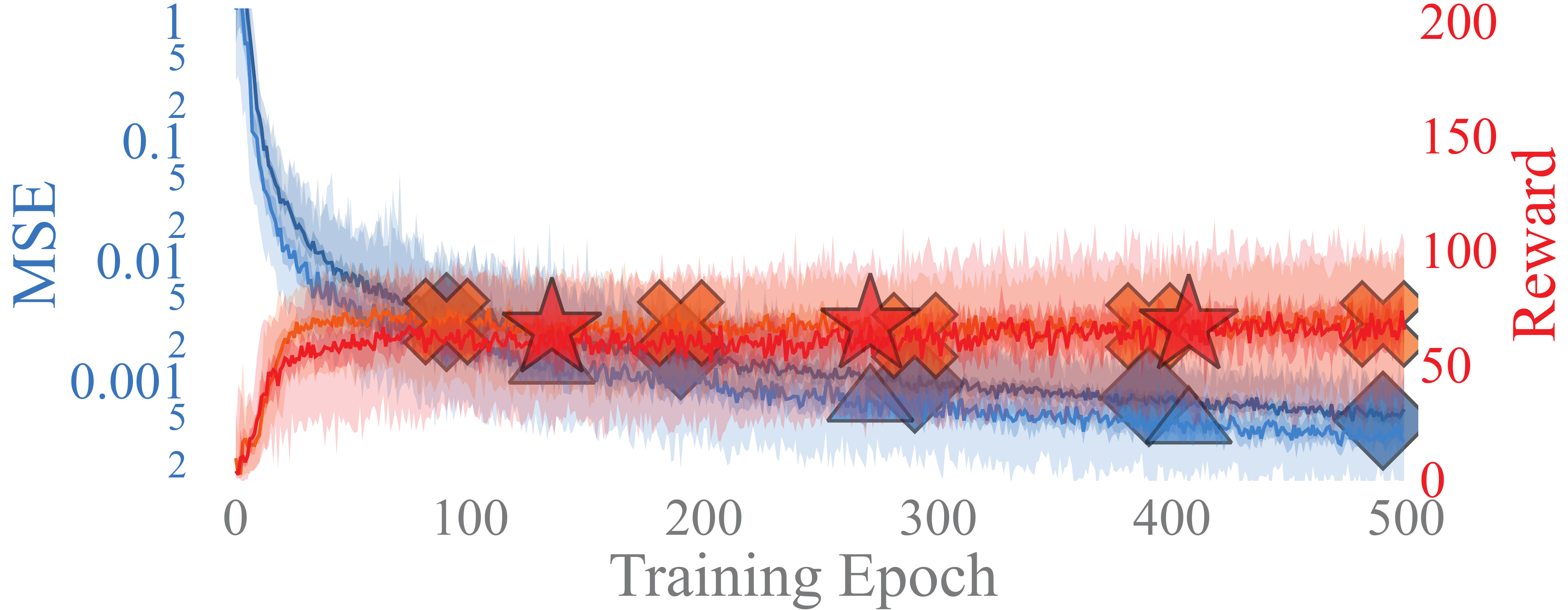}}
      }
\end{figure}

\begin{figure}[t]
   \floatconts{fig:cp-tc-eval}
   {
   \vspace{-10pt}
   \caption{The effect of the dataset choice on model ($P$) training and accuracy in different regions of the state-space, $N=50$ per model type. (\textit{Left}) when training on the complete dataset, the model begins over-fitting to the on-policy data even before the performance drops in the controller. (\textit{Right}) A model trained only on policy data does not accurately model the entire state-space. 
   The validation loss is still improving slowly at 500 epochs in both scenarios.
    \vspace{-5pt}
    }
    }
   {
    \centering{ \small{\cblock{31}{119}{180} Grid Data (\textcolor[rgb]{.12,.46,.71}{$	\bigtriangleup$})\quad
    \cblock{23}{190}{207} Policy Data (\textcolor[rgb]{.09,.73,.81}{$\bigcirc$})\quad
    \cblock{44}{160}{44} Expert Data (\textcolor[rgb]{.17,.63,.17}{$\Diamond$})\quad
    \cblock{255}{69}{46} Episode Reward (\textcolor[rgb]{1,.27,.18}{$\mathbb{X}$}) 
    }} \\ %
    \subfigure[\textbf{Trained:} grid\quad \textbf{Tested:} expert, on-policy]{ 
      \includegraphics[width=0.48\linewidth]{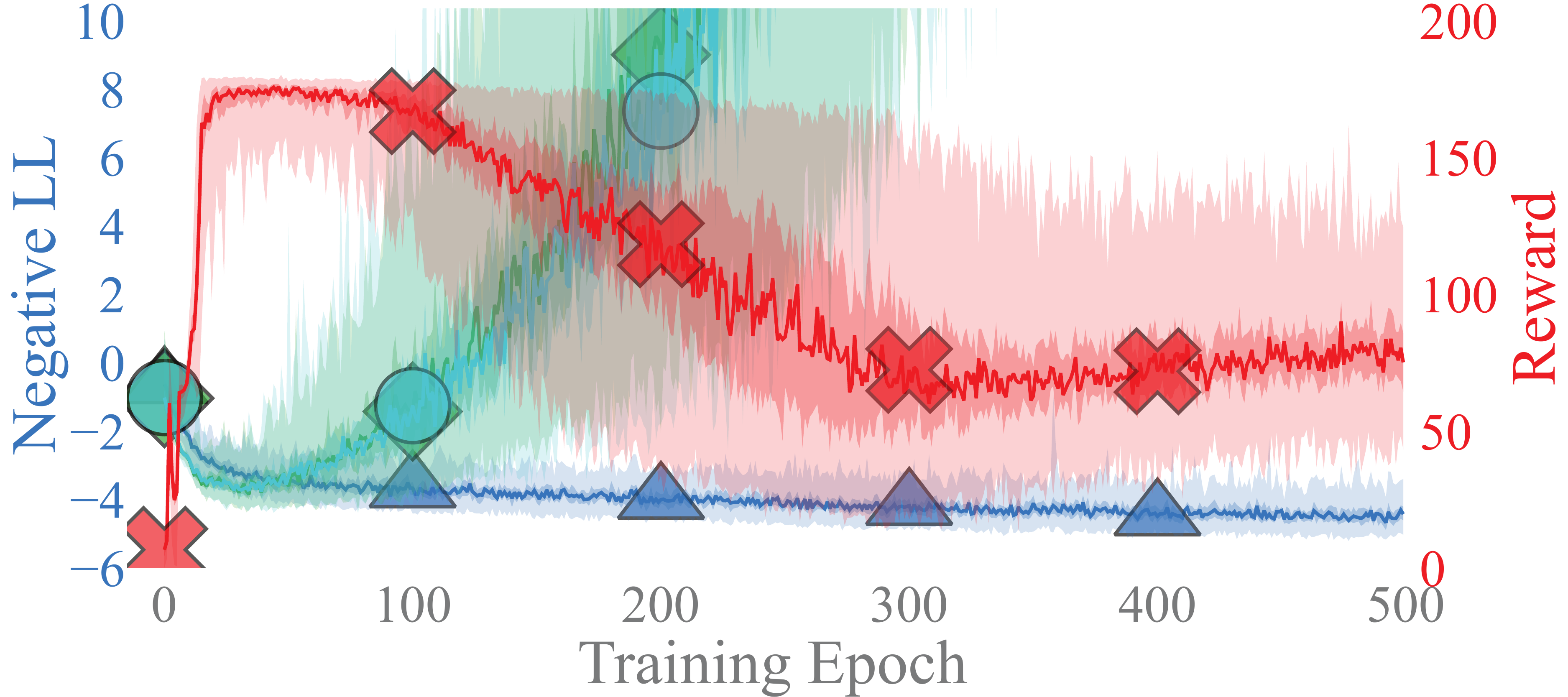}}%
    \hfill
    \subfigure[\textbf{Trained:} on-policy\quad \textbf{Tested:} expert, grid]{ 
      \includegraphics[width=0.48\linewidth]{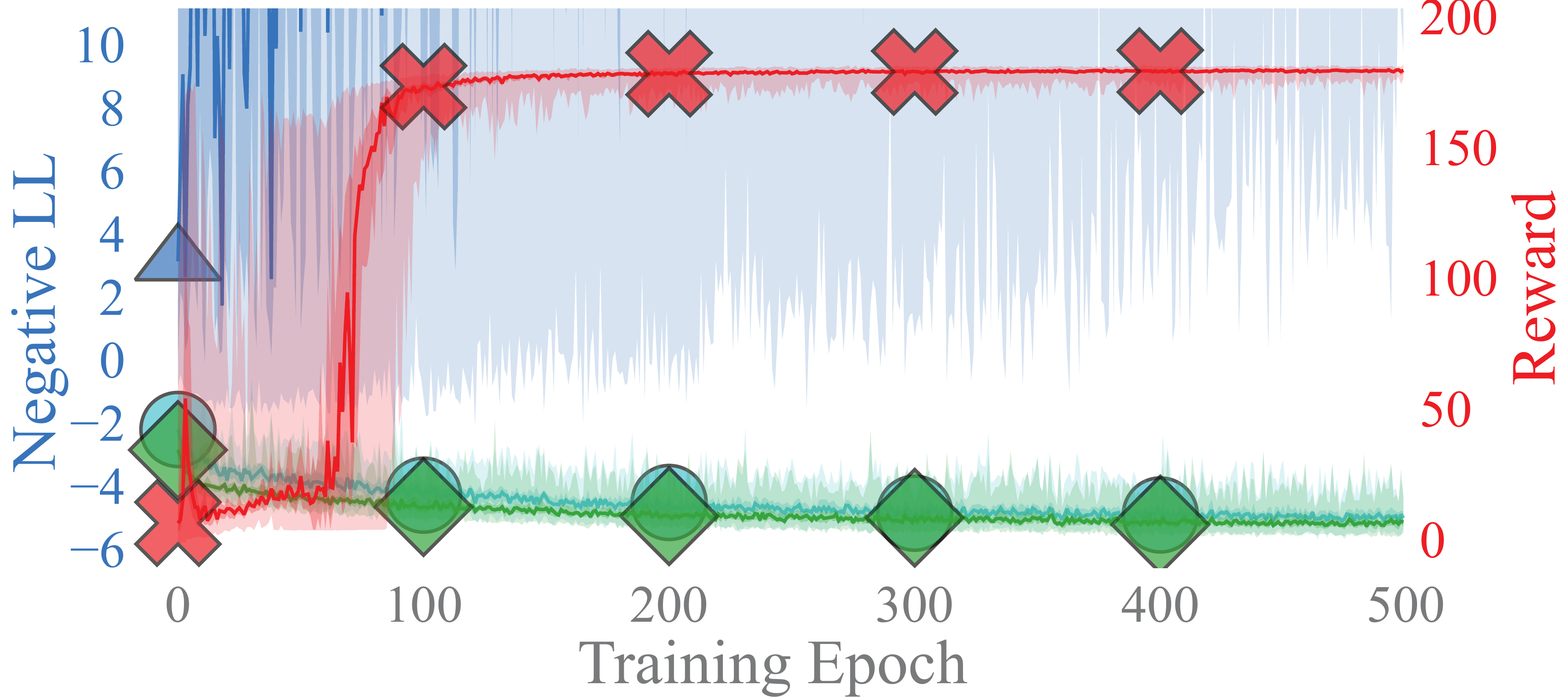}}
      }

\end{figure}

This section explores how model training impacts performance at the per-epoch level.
These experiments shed light onto the impact of the strong model assumptions outlined in \sect{sec:origin}.
As a dynamics model is trained, there are two key inflection points - the first is the training epoch where episode reward is maximized, and the second is when error on the validation set is optimized.
These experiments highlight the disconnect between three practices in MBRL a) the assumption that the on-policy dynamics data can express large portions of the state-space, b) the idea that simple neural networks can satisfactorily capture complex dynamics, c) and the practice that model training is a simple optimization problem disconnected from reward.
Note that in the figures of this section we use Negative Log-Likelihood (NLL) instead of LL, to reduce visual clutter.  

For the grid cartpole dataset, \fig{fig:cp-tc} shows that the reward is maximized at a drastically different time than when validation loss is minimized for $P$, $PE$ models. 
\fig{fig:cp-tc-eval} highlights how the trained models are able to represent other datasets than they are trained on (with additional validation errors). 
\fig{fig:cp-tc-eval}b shows that on-policy data will not lead to a complete dynamics understanding because the grid validation data rapidly diverges. 
When training on grid data, the fact that the on-policy data diverges in \fig{fig:cp-tc-eval}a before the reward decreases is encouraging as objective mismatch may be preventable in simple tasks. 
Similar experiments on half cheetah are omitted because models for this environment are trained incrementally on aggregated data rather then fully on each dataset~(\citet{chua2018deep}).

\subsection{Decoupling Model Loss from Controller Performance}
\label{sec:control}

We now study how differences in dynamics models -- \textit{even if they have similar LLs} -- are reflected in control policies to show that a accurate dynamics model does not guarantee performance.

\begin{figure*}[t]
    \centering
    \includegraphics[width=.8\textwidth]{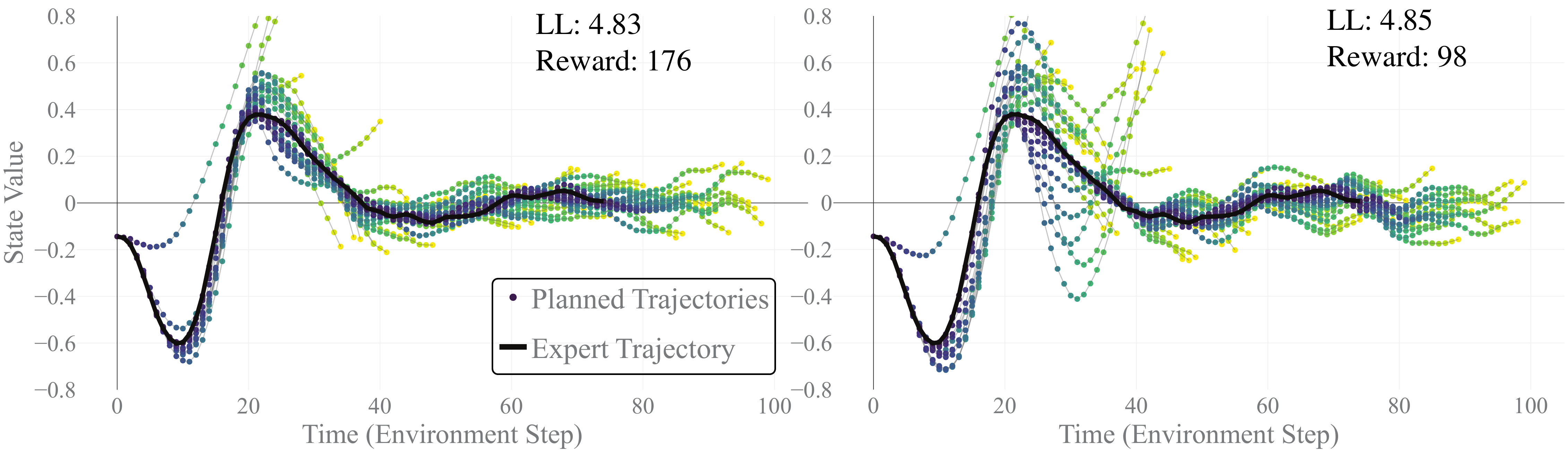}
    \vspace{-10pt}
    \caption{Example of planned trajectories along the expert trajectory for (\textit{left}) a learned model and (\textit{right}) the adversarially generated model trained to lower the reward. The planned control sequences are qualitatively similar except for the peak at $t=25$.
    There, the adversarial attack applies a small nudge to the dynamics model parameters that significantly influences the control outcome with minimal change in terms of LL.
    }
    \label{fig:cp-control}
\end{figure*}

\begin{wrapfigure}{r}{.45\textwidth}
    \centering
    \vspace{-10pt}
    \includegraphics[width=\linewidth]{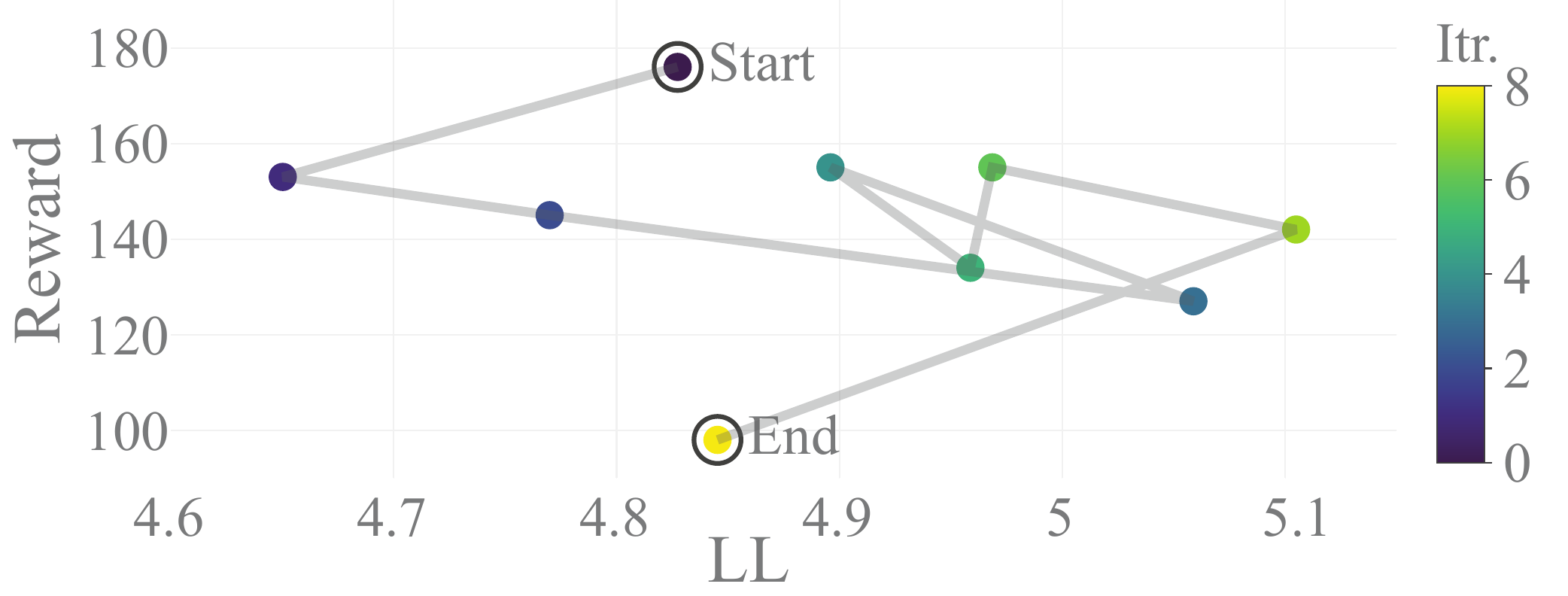}
    \caption{Convergence of the CMA-ES population's best member.
    }
    \label{fig:cma}
    \vspace{-10pt}
\end{wrapfigure}
\paragraph{Adversarial attack on model performance}
We performed an adversarial attack~\citep{szegedy2013intriguing} on a deep dynamics model so that it attains a high likelihood but low reward. 
Specifically, we fine-tune the deep dynamics model's last layer with a zeroth-order optimizer, CMA-ES, (the cumulative reward is non-differentiable) to lower reward with a large penalty if the  validation likelihood drops.
As a starting point for this experiment we sampled a $P$ dynamics model from the last trial of a PETS run on cartpole.
This model achieves reward of $176$ and has a LL of $4.827$ on it's
on-policy validation dataset.
Using CMA-ES, we reduced the on-policy reward of the model to $98$, on 5 trials,
while slightly improving the LL; the CMA-ES convergence is shown in \fig{fig:cma} and the difference between the two models is visualized in \fig{fig:cp-control}.
Fine tuning of all model parameters would be more likely to find sub-optimal performing controllers because the output layer consists of about $1\%$ of the total parameters.
This experiment shows that the model parameters that achieve a low model loss inhabit a broader space than the subset that also achieves high reward.


\section{Addressing Objective Mismatch During Model Training}
\label{sec:solution}

	\begin{figure*}[t]
    \centering
    \includegraphics[width=1\textwidth]{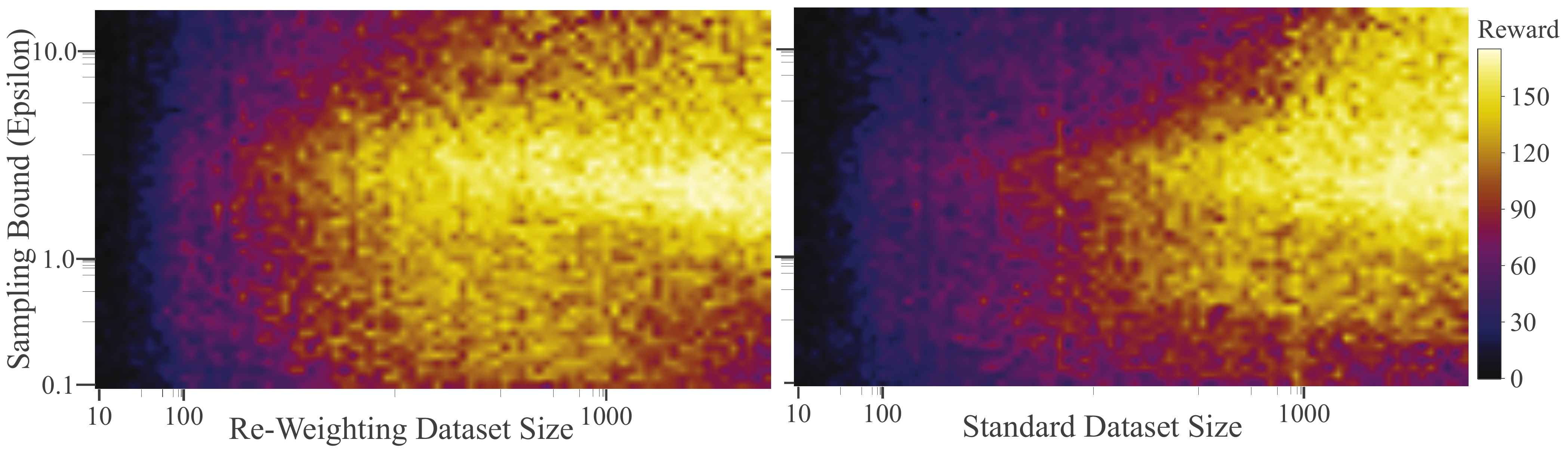}
    \caption{Mean reward of PETS trials ($N_{trials}=100$), with (\textit{left}) and without (\textit{right}) model re-weighting, on a log-grid of dynamics model training sets with number of points $\mathrm{S} \in [10,2500]$ and sampling optimal-distance bounds $\epsilon \in [.28, 15.66]$.  
    The re-weighting improves performance for smaller dataset sizes, but suffers from increased variance in larger set sizes.
    The performance of PETS declines when the dynamics model is trained on points too near to the optimal trajectory because the model lacks robustness when running online with the stochastic MPC.
    }
    \label{fig:heatmap}
        \vspace{-10pt}
\end{figure*}

Tweaking dynamics model training can partially mitigate the problem of objective mismatch.
Taking inspiration from imitation learning, we propose that the learning capacity of the model would be most useful when accurately modeling the dynamics along trajectories that are relevant for the task at hand, while maintaining knowledge of nearby transitions for robustness under a stochastic controller.
Intuitively, it is more important to model accurately the dynamics along the optimal trajectory, rather than modeling part of the state-action space that might never be visited to solve the task.
For this reason, we now propose a model loss aimed at alleviating this issue.

\begin{wrapfigure}{r}{.36\textwidth}
    \centering
    \vspace{-10pt}
    \includegraphics[width=\linewidth]{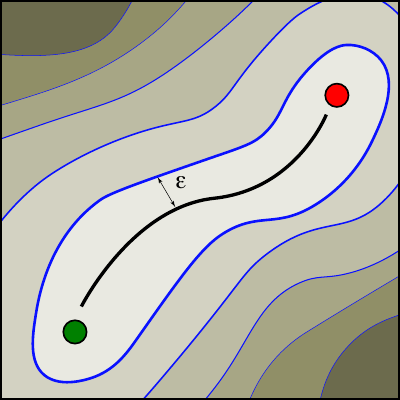}
    \vspace{-13pt}
        \caption{We propose to re-weight the loss of the dynamics model w.r.t. the distance~$\epsilon$ from the optimal trajectory. 
    }
    \label{fig:weightmot}
    \vspace{-5pt}
\end{wrapfigure}
Given a element of a state space $(s_i, a_i)$, we quantify the distance of any two tuples, $d_{i,j}$. 
With this distance, we re-weight the loss, $l(y)$, of points further from the optimal policy to be lower, so that points in the optimal trajectory get a weight $\omega(y)=1$, and points at the edge of the grid dataset used in \sect{sec:experiments} get a weight $\omega(y)=0$. 
Using the expert dataset discussed in \sect{sec:experiments} as a distance baseline, we generated $25e6$ tuples of $(s, a, s')$ by uniformly sampling across the state-action space of cartpole.
We sorted this data by taking the minimum orthogonal distance, $d^*$, from each of the points to the $200$ elements in the expert trajectory.
To create different datasets that range from near-optimal to near-global, we vary the distance bound~$\epsilon$, and number of training points, $\mathrm{S}$.
This simple form of re-weighting the neural network loss
, shown in \eq{eq:weight},  demonstrated an improvement in sample efficiency to learn the cartpole task, as seen in \fig{fig:heatmap}. 
Unfortunately, this approach is impractical when the optimal trajectory is not known in advance. 
However, future work could develop an iterative method to jointly estimate and re-weight samples in an online training method to address objective mismatch. 
\begin{subequations}\label{eq:4}
\begin{gather}
    \textbf{Weighting} \ \ \omega(y) = ce^{-d^*(y)} \quad \quad  
    \textbf{Standard} \ \ l(\hat{y},y) \quad \quad \textbf{Re-weight} \ \   l(\hat{y},y)\cdot \omega(y)  \tag{\theequation a,b,c} 
\label{eq:weight}
\end{gather}
\end{subequations}


\section{Discussion, Related Work, and Future Work}
\label{sec:Discussion}

\emph{Objective mismatch} impacts the performance of MBRL -- our experiments have gone deeper into this fragility. 
Beyond the re-weighting of the LL presented in \sect{sec:solution}, here we summarize and discuss other relevant works in the community.

\paragraph{Learning the dynamics model to optimize the task performance}
Most relevant are research directions on controllers that directly connect the reward signal back to the controller.
In theory, this exactly solves the model mismatch problem, but in practice the current approaches have proven difficult to scale to complex systems.
One way to do this is by designing systems that are fully differentiable and backpropagating the task reward through the dynamics.
This has been investigated with differentiable MPC~\citep{amos2018differentiable} and Path Integral control~\citep{okada2017path}, 
Universal Planning Networks~\citep{srinivas2018universal} propose a differentiable planner that unrolls gradient descent steps over the action space of a planning network.
\citet{Bansal2017} use a zero-order optimizer to maximize the controller's performance without having to compute gradients explicitly.

\paragraph{Add heuristics to the dynamics model structure or training process to make control easier}
If it is infeasible or intractable to shape the dynamics of a controller, an alternative is to add heuristics to the training process of the dynamics model.
These heuristics can manifest in the form of learning a latent space that is locally linear, \eg, in
Embed to Control and related methods~\citep{watter2015embed}, 
by enforcing that the model makes long-horizon predictions~\citep{ke2019learning}, ignoring uncontrollable parts of the state space~\citep{ghosh2018learning}, detecting and correcting when a predictive model steps off the manifold of reasonable states~\citep{talvitie2017self},
adding reward signal prediction on top of the latent space~\cite{gelada2019deepmdp}, or adding noise when training transitions~\cite{mankowitz2019robust}.
\citet{farahmand2017value, farahmand2018iterative} also attempts to re-frame the transitions to incorporate a notion of the downstream decision or reward.
Finally, \citet{Singh2019Learning} proposes stabilizability constraints to regularize the model and improve the control performance.
None of these paper formalize or explore the underlying mismatch issue in detail.

\paragraph{Continuing Experiments} Our experiments represent an initial exploration into the challenges
of objective mismatch in MBRL. 
\sect{sec:traincollect} is limited to cartpole due to computational challenges of training with large dynamics datasets and \sect{sec:control} could be strengthened by defining quantitative comparisons in controller performance.
Additionally, these effects should be quantified in other MBRL algorithms such as MBPO~\citep{janner2019trust} and POPLIN~\citep{wang2019exploring}.


\section{Conclusion}
\label{sec:conclusion}

	This paper identifies, formalizes and analyzes the issue of objective mismatch in MBRL.
This fundamental disconnect between the likelihood of the dynamics model, and the overall task reward emerges from incorrect assumptions at the origins of MBRL.
Experimental results highlight the negative effects that objective mismatch has on the performance of a current state-of-the-art MBRL algorithm.
In providing a first insight on the issue of objective mismatch in MBRL, we hope future work will deeply examine this issue to overcome it with a new generation of MBRL algorithms.


\acks{We thank Rowan McAllister and Kurtland Chua for useful discussions.}

\clearpage

\bibliography{main}

\clearpage
\appendix
\section{Effect of Dataset Distribution when Learning }

\begin{wrapfigure}{r}{.5\textwidth}
    \centering
    \vspace{-15pt}
    \small{
    Number of random transitions: \\ \cblock{31}{119}{180} 200, \quad
    \cblock{255}{69}{46} 2000, \quad 
    \cblock{44}{160}{44} 4000, \quad
    \cblock{205}{100}{46} 20000 }
    \includegraphics[width=\linewidth]{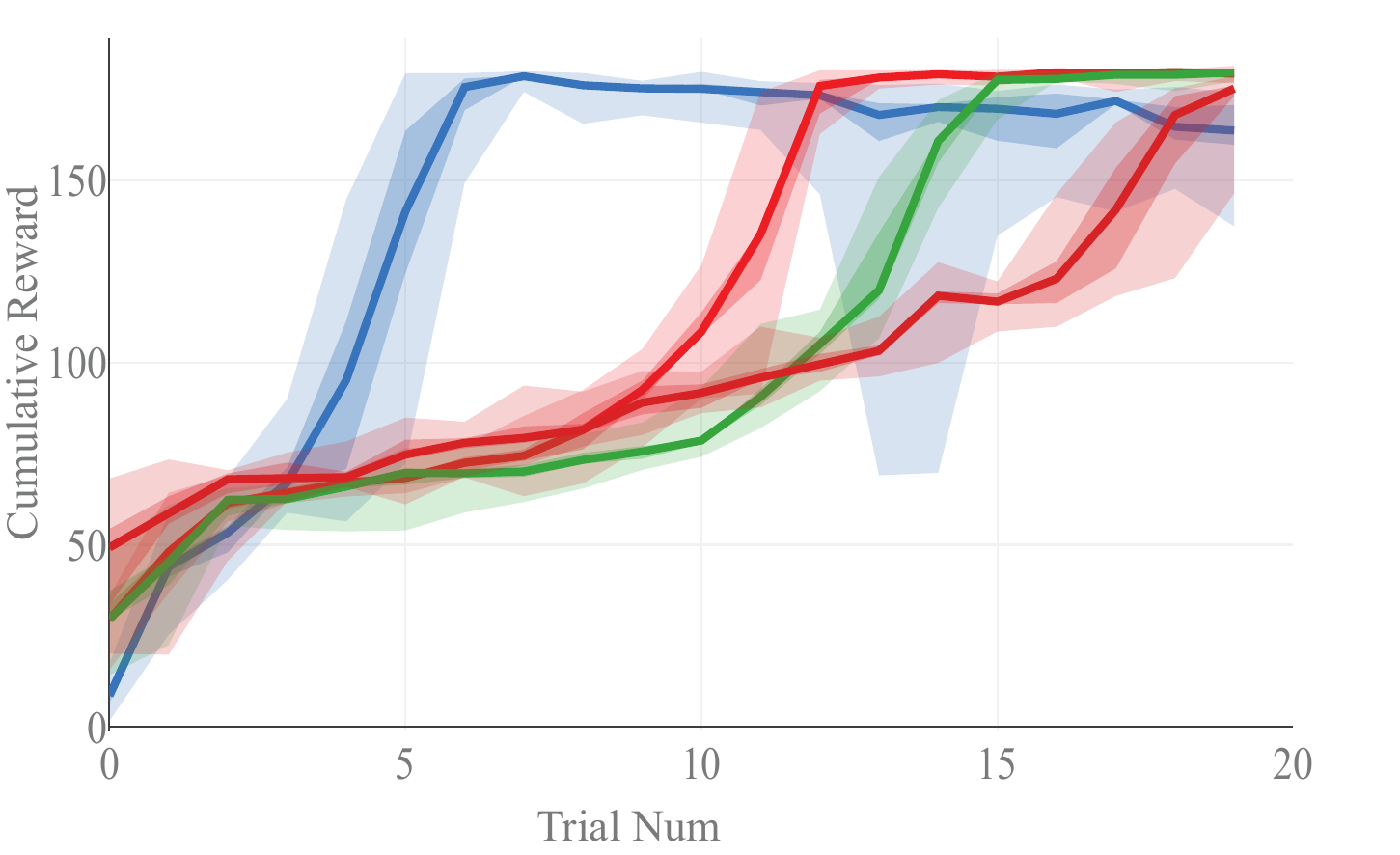}
    \caption{Cartpole (Mujoco simulations) learning efficiency is suppressed when additional data not relevant to the task is added to the dynamics model training set. 
    This effect is related to the issue of objective mismatch because model training needs to account for potential off-task data.}
    \label{fig:bubbling}
    \vspace{-25pt}
\end{wrapfigure}
Learning speed can be slowed by many factors in dataset distribution, such as adding additional irrelevant transitions. When extra transitions from a specific area of the state space are included in the training set, the dynamics model will spend increased expression on these transitions. LL of the model will be biased down as it learns this data, but it will reduce the learning speed as new, more relevant transitions are added to the training set.

Running cartpole random data collection with a short horizon of 10 steps (while forcing initial babbling state to always be 0), for 20, 200,400 and 2000 babbling roll-outs (that sums up to 200, 2000, 4000 and 20000 transitions in the dataset finally shows some regression in the learning speed for runs with more useless data in the motor babbling. This data highlights the importance of careful exploration vs exploitation trade-offs, or changing how models are trained to be selective with data.

\section{Task Generalization in Simple Environments}
\begin{wrapfigure}{r}{.5\textwidth}
\centering
    \small{
    \cblock{31}{119}{180} Validation Error \quad
    \cblock{255}{69}{46} Episode Reward  \\
    }
    \includegraphics[width=.5\textwidth]{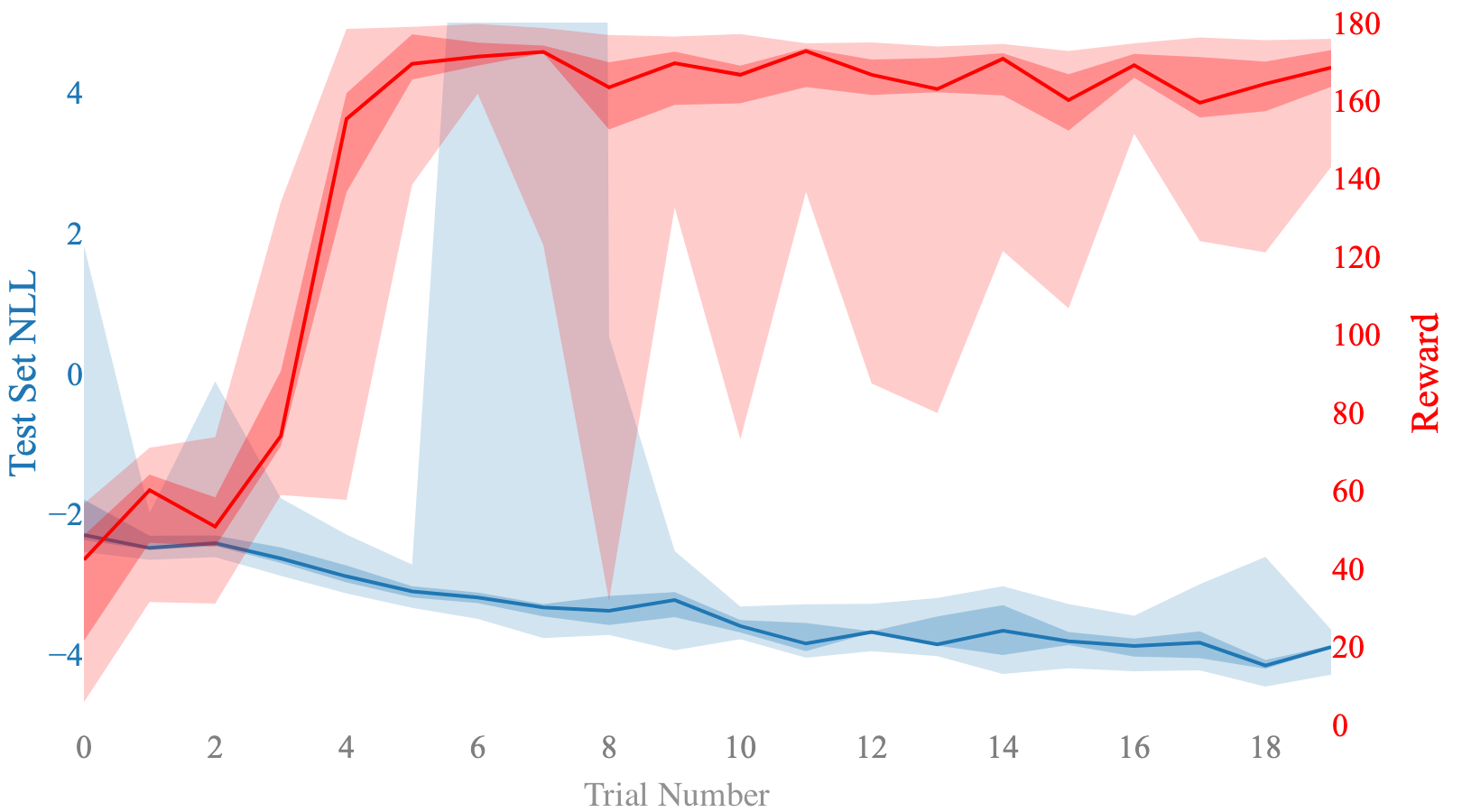}
        \vspace{-15pt}
    \caption{Learning curve for the standard Cartpole task used in this paper ($X_{goal}=0$). 
    The median reward from 10 trials is plotted with the mean NLL of the dynamics models at each iteration. 
    The reward reaches maximum (180) well before the NLL is at it's minimum.
    }
    \label{fig:general_base}
    \vspace{-25pt}
    
\end{wrapfigure}
\begin{figure}[t]
   \floatconts
      {fig:general}
      {\vspace{-15pt} \centering
    
    \cblock{31}{119}{180} $|X_{goal}| = .1$ \quad
    \cblock{0}{255}{255}  $|X_{goal}| = .2$ \quad
    \cblock{128}{128}{0}  $|X_{goal}| = .5$ \quad
    \cblock{128}{128}{128}  $|X_{goal}| = 1.0$  \\ 
    \caption{MPC control with different reward functions with the same dynamics models loaded from 
    trials shown in \fig{fig:general_base}.
    The cartpole solves tasks further from $0$ proportional to the state space coverage (\textit{Goal further from zero causes reduced performance}). 
    The distribution of $x$ data encountered is shown in \fig{fig:general_dist}. }}
      {%
        \subfigure[$X_{goal} < 0$]{\label{fig:left}%
          \includegraphics[width=0.48\linewidth]{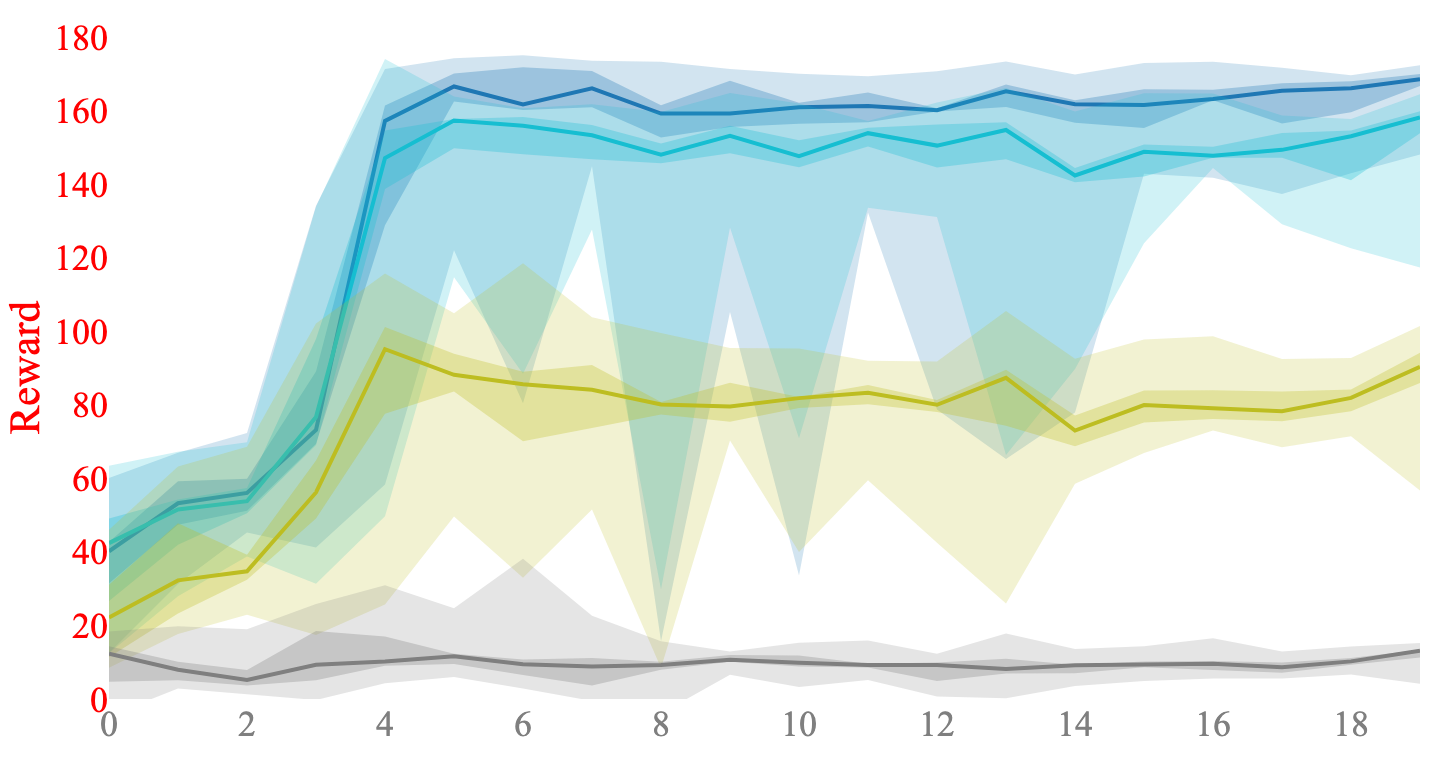}}%
        \hfill
        \subfigure[$X_{goal} > 0$]{\label{fig:right}%
          \includegraphics[width=0.48\linewidth]{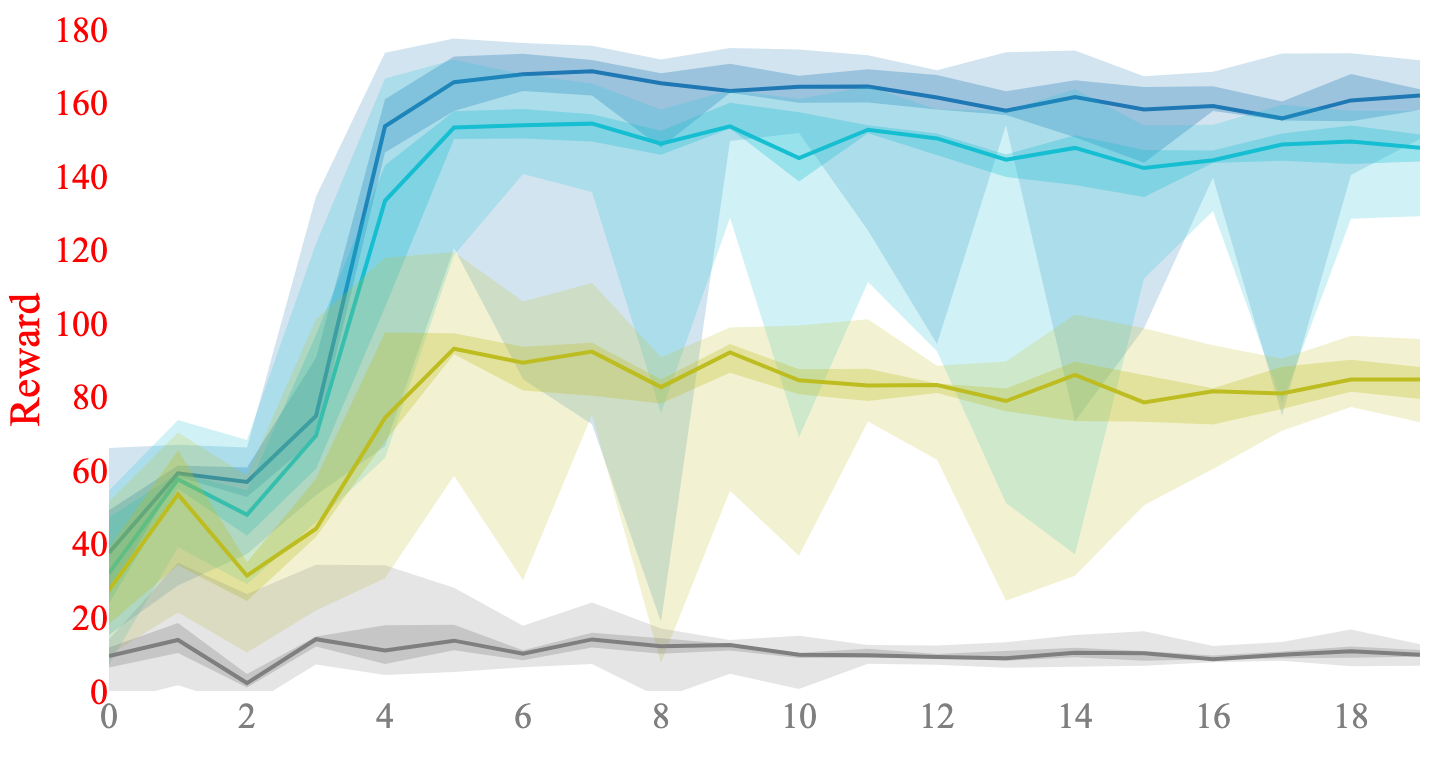}}
      }
    \vspace{-10pt}
\end{figure}

In this section, we compare the performance of a model trained on data for the standard cartpole task (x position goal at 0) to policies attempting to move the cart to different positions in the x-axis. 
\fig{fig:general_base} is a learning curve of PETS with a PE model using the CEM optimizer. 
Even though performance levels out, the NLL continues to decrease as the dynamics models accrue more data. 
With more complicated systems, such as halfcheetah, the reward of different tasks verses global likelihood of the model would likely be more interesting (especially with incremental model training) - we will investigate this in future work. 
Below, we show that the dynamics model generalizes well to tasks close to zero (both positive in (\fig{fig:right}) and negative positions (\fig{fig:left}), but performance drops off in areas the training set does not cover as well.

Below the learning curves in \fig{fig:general_dist}, we include snapshots of the distributions of training data used for these models at different trials, showing how coverage relates to reward in cartpole. It is worth investigating how many points can be removed from the training set while maintaining peak performance on each task.

\begin{figure}[t]
   \floatconts
      {fig:general_dist}
      {\vspace{-15pt}
    \caption{Distribution of $x$ position encountered during the trials shown in \fig{fig:general_base}. 
    The distribution converges to a high concentration around $0$, making it difficult for MPC to control outside of the area close to $0$. }}
      {%
        \subfigure[$x$ distribution after trial $1$.]{ 
          \includegraphics[width=0.32\linewidth]{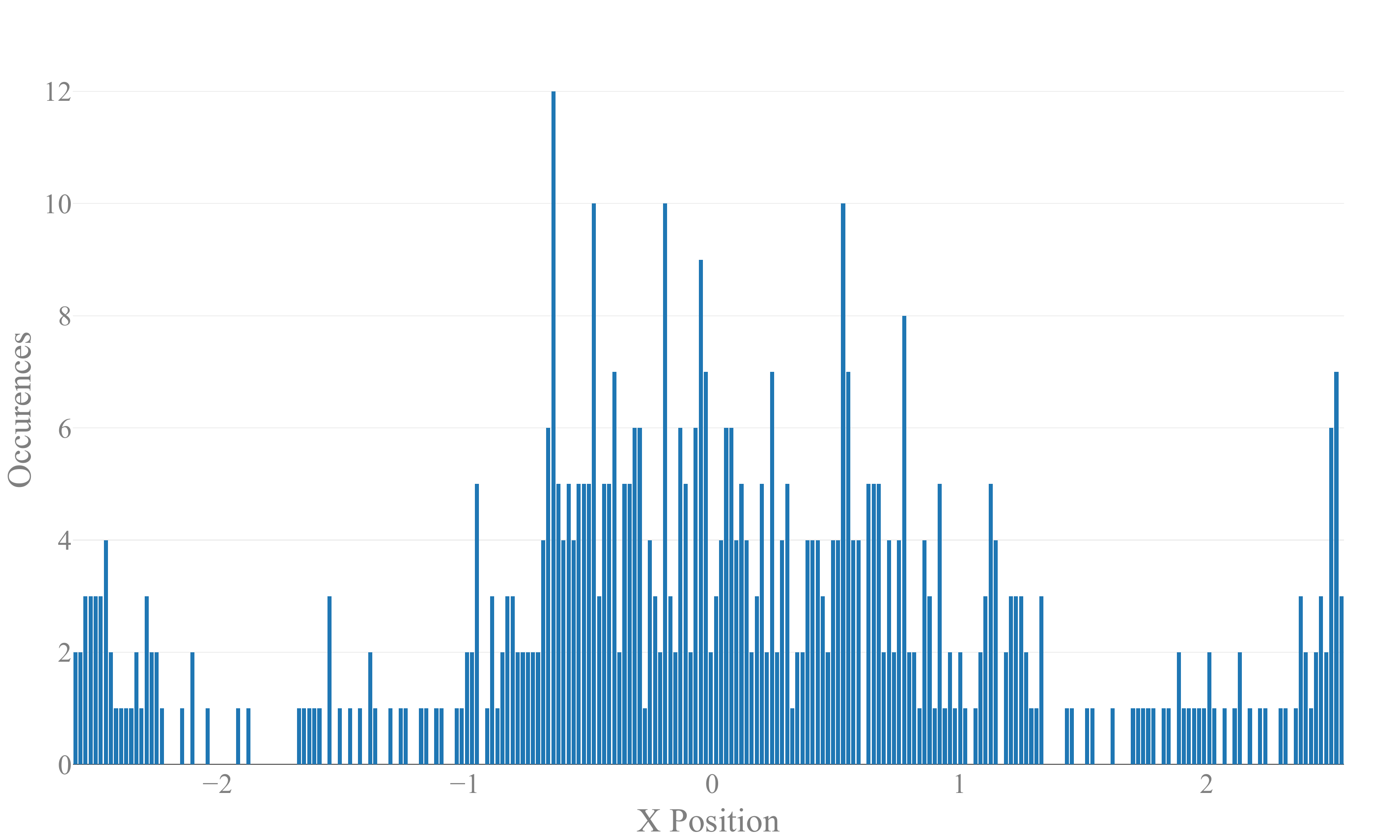}}%
        \subfigure[$x$ distribution after trial $5$.]{ 
          \includegraphics[width=0.32\linewidth]{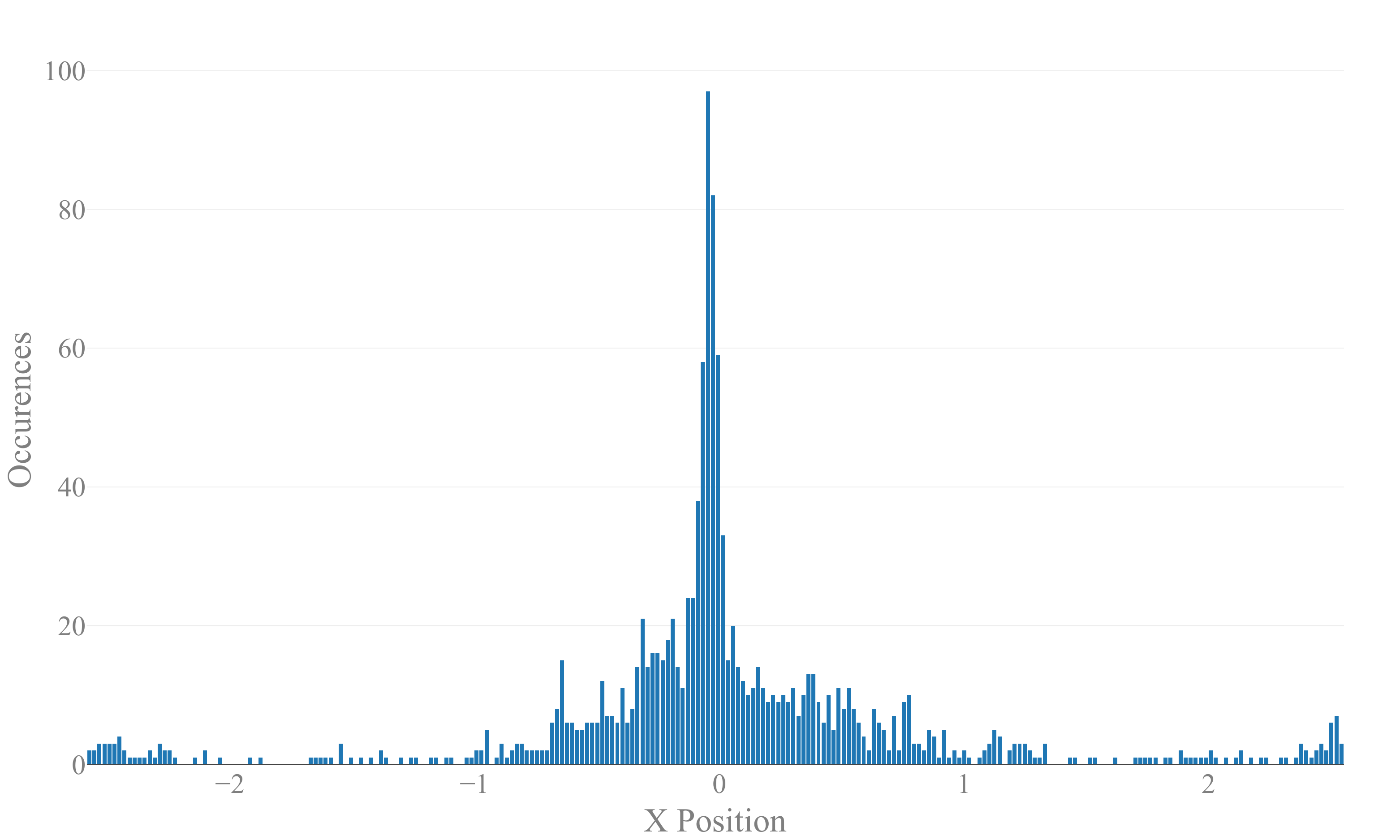}} %
      \subfigure[$x$ distribution after trial $20$.]{ 
      \includegraphics[width=0.33\linewidth]{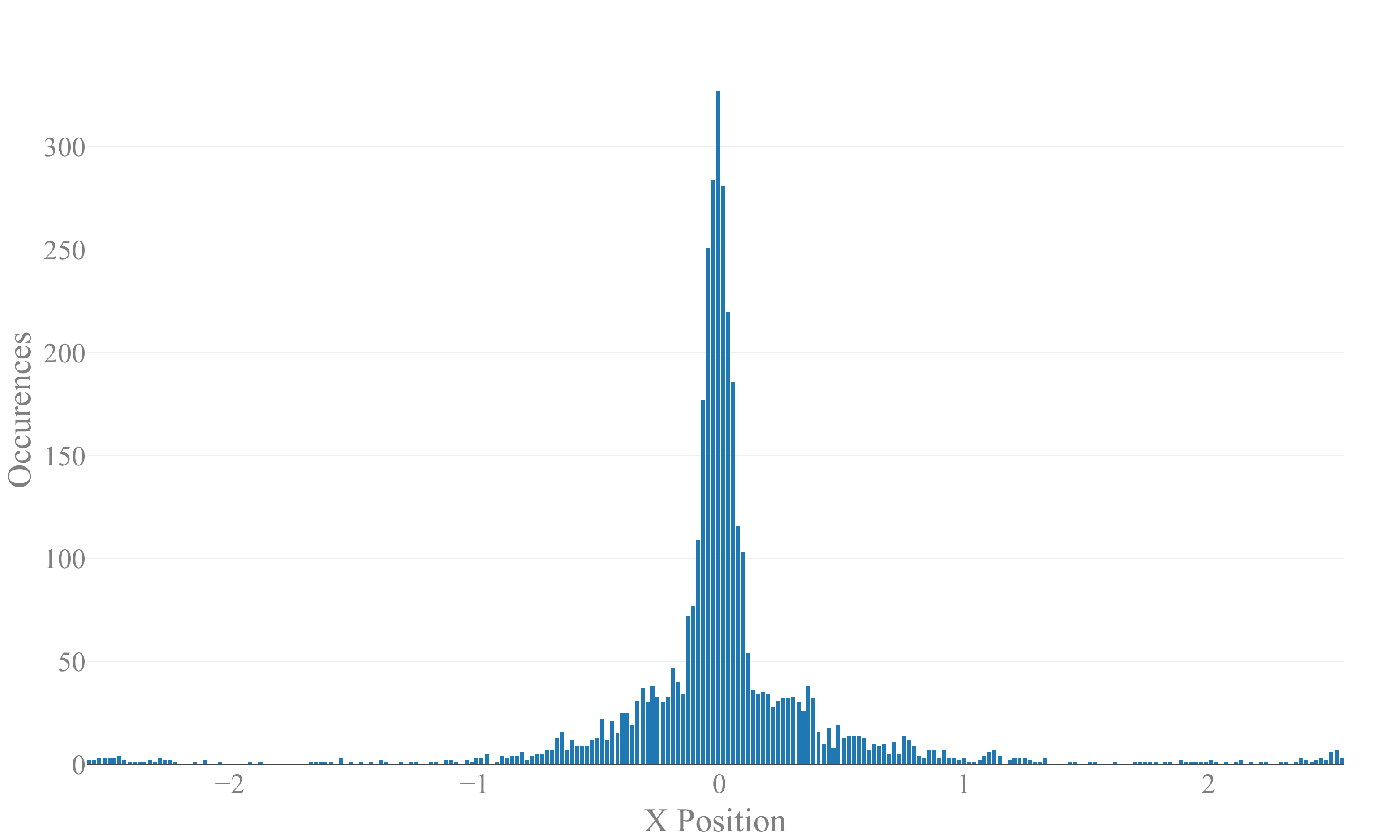}}
      }
    \vspace{-10pt}
\end{figure}

\section{Validating models with trajectories rather than random tuples}
The goal of the dynamics model for planning is to be able to  predict stable
long term roll-outs conditioned on different actions. 
This is because in sampling based control, the MPC chooses the best planned trajectory, not the best collection of random one-step predictions (akin to random, small batches).
Different results could be found in short, simulated approaches such as~\cite{janner2019trust}, where predictive accuracy is validated under policy shift for one-step predictions.
We propose that evaluating the test set when training a dynamics model could be 
more reliable (in terms of relation between loss and reward under the induced planning-based controller) if the model is validated
on batches consisting entirely of the same trajectory, rather then a random shuffle of points.
When randomly shuffling points, the test loss can be easily dominated by an outlier in each batch.

\begin{figure}[t]
  \floatconts
      {fig:traj-loss}
      {\vspace{-20pt} \centering
    \caption{There is a slight increase in the correlation between LL and reward when training on cartpole trajectories rather than random samples. 
    This could be one small step in the right direction of solving objective mismatch.
      }}
      {%
        \subfigure[CP LL from trajectory based loss ($\rho = .36$).]{\label{fig:cpleft}%
          \includegraphics[width=0.48\linewidth]{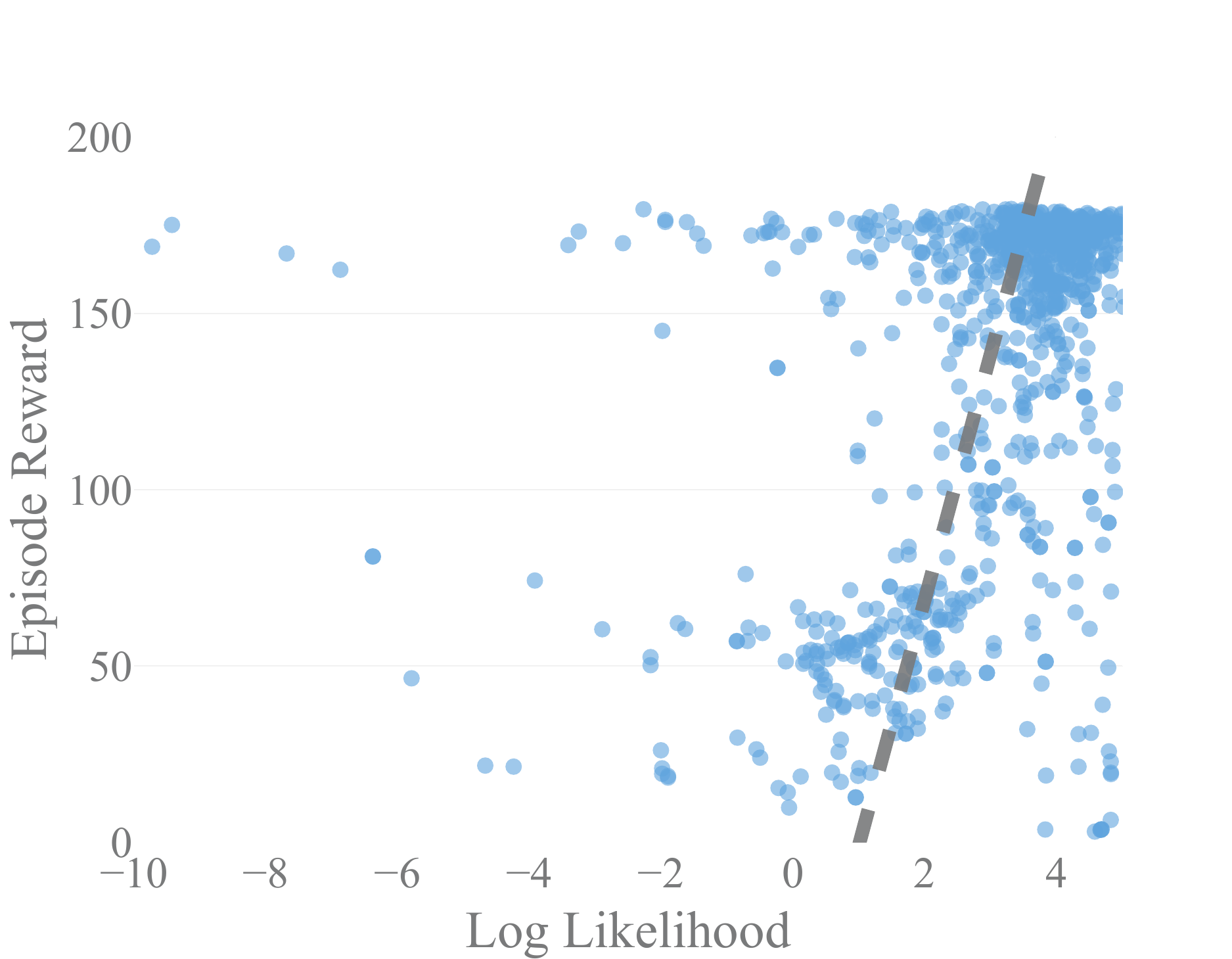}}%
        \hfill
        \subfigure[CP LL for standard loss formulation ($\rho = .34$)]{\label{fig:cpright}%
          \includegraphics[width=0.48\linewidth]{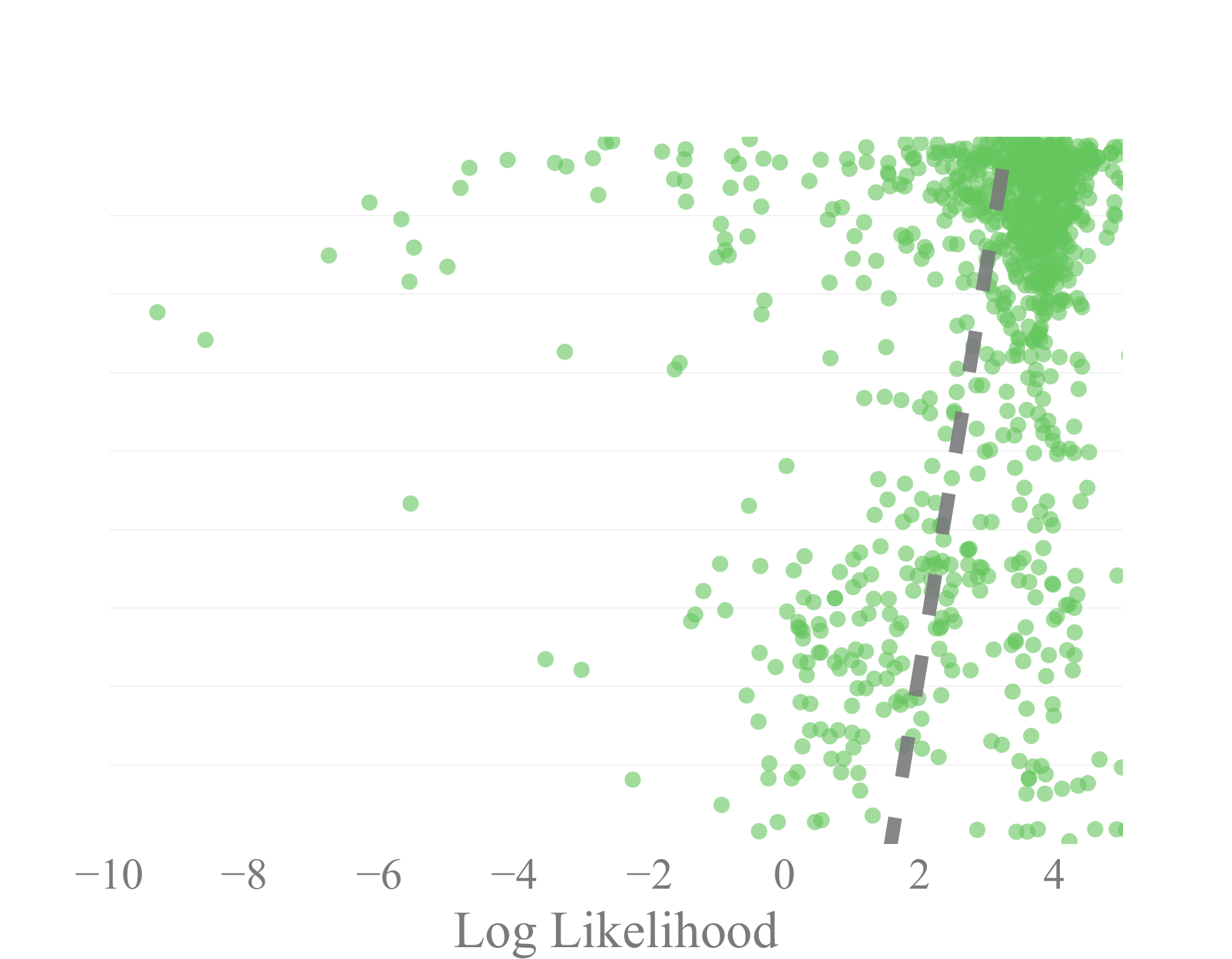}}
      }
    \vspace{-10pt}
\end{figure}

\begin{figure}[t]
   \floatconts
      {fig:nllVr-traj}
      {\caption{Validation of model LL versus reward with different types of validation of the half cheetah models. 
      (\textit{left}) Is a new method for training, where each batch of the validation set is a complete subsection of a trajectory in the aggregated dataset.
      (\textit{center}) We compare the trajectory loss to the regularization that would be provided when just validating with larger batches, which would reduce variance from outliers.
      (\textit{right}) Copied from figure \fig{fig:nllVr}e where validation is done on small batches randomly sampled.
      }}
      {%
        \subfigure[HC traj. loss ($\rho=0.63$)]{\label{fig:hc-sub-traj}%
          \includegraphics[width=0.32\linewidth]{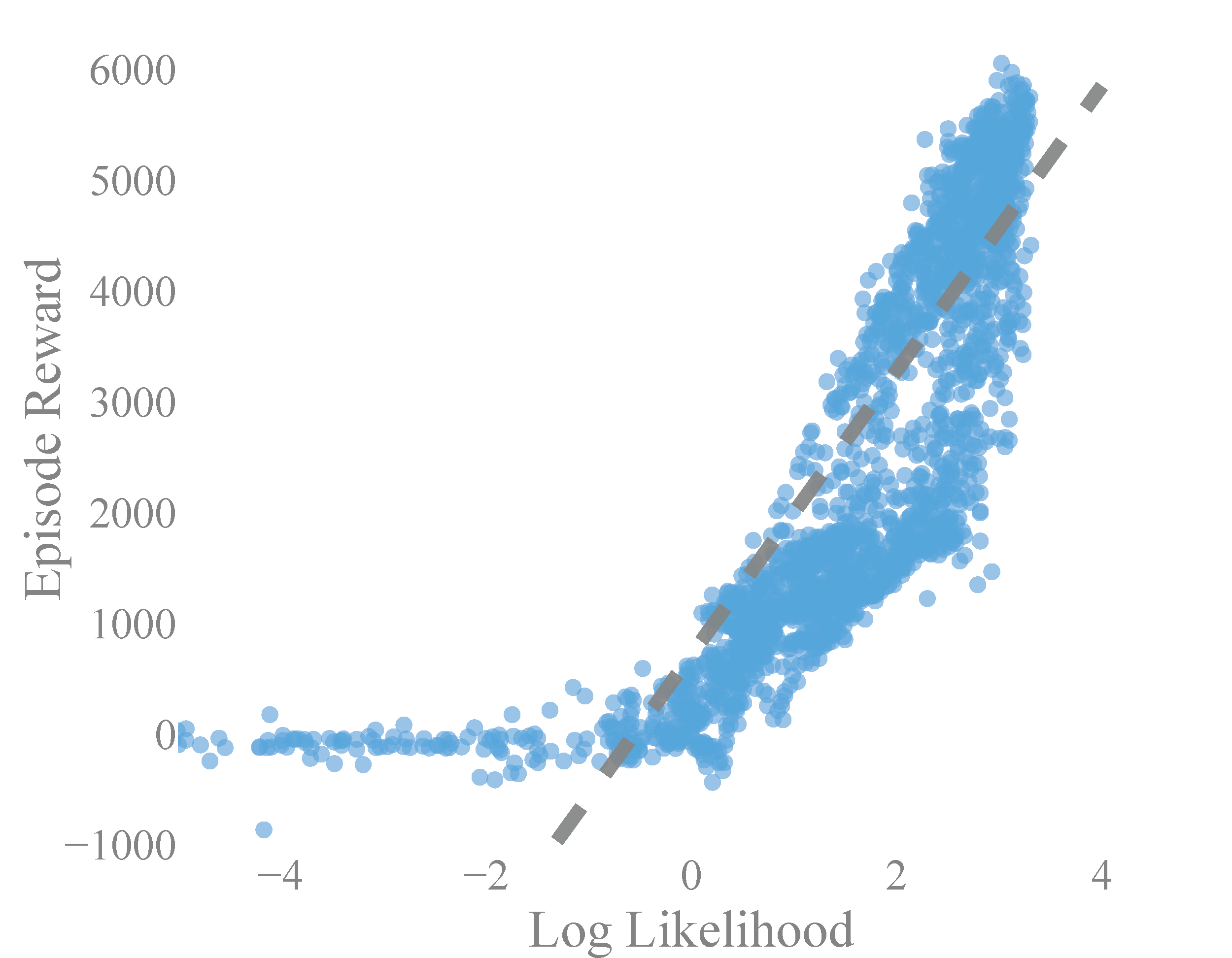}}
          \subfigure[HC large batch ($\rho=0.54$)]{\label{fig:hc-sub-batch}%
          \includegraphics[width=0.32\linewidth]{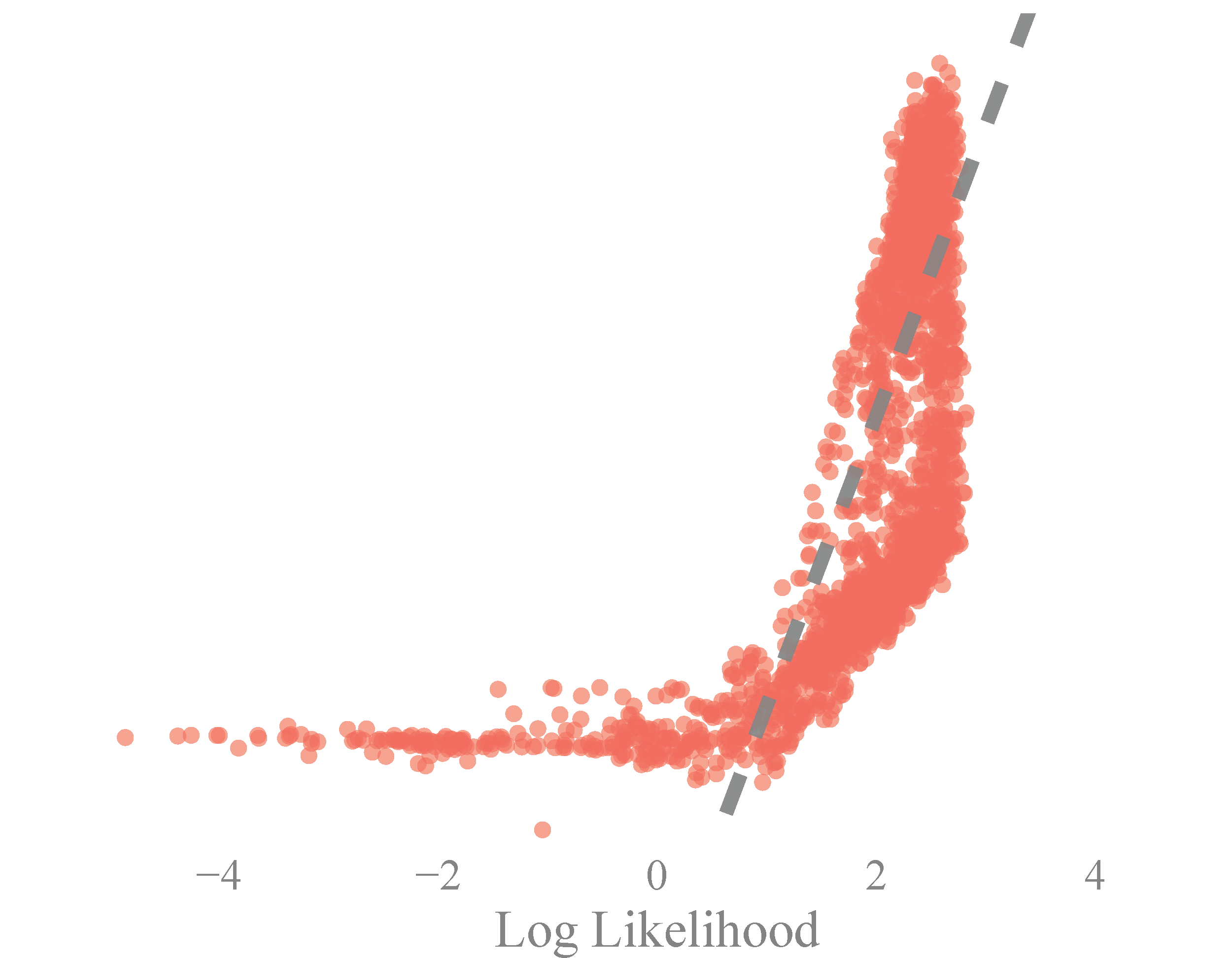}}
          \subfigure[HC on-policy ($\rho=0.49$)]{\label{fig:hc-sub}%
          \includegraphics[width=0.32\linewidth]{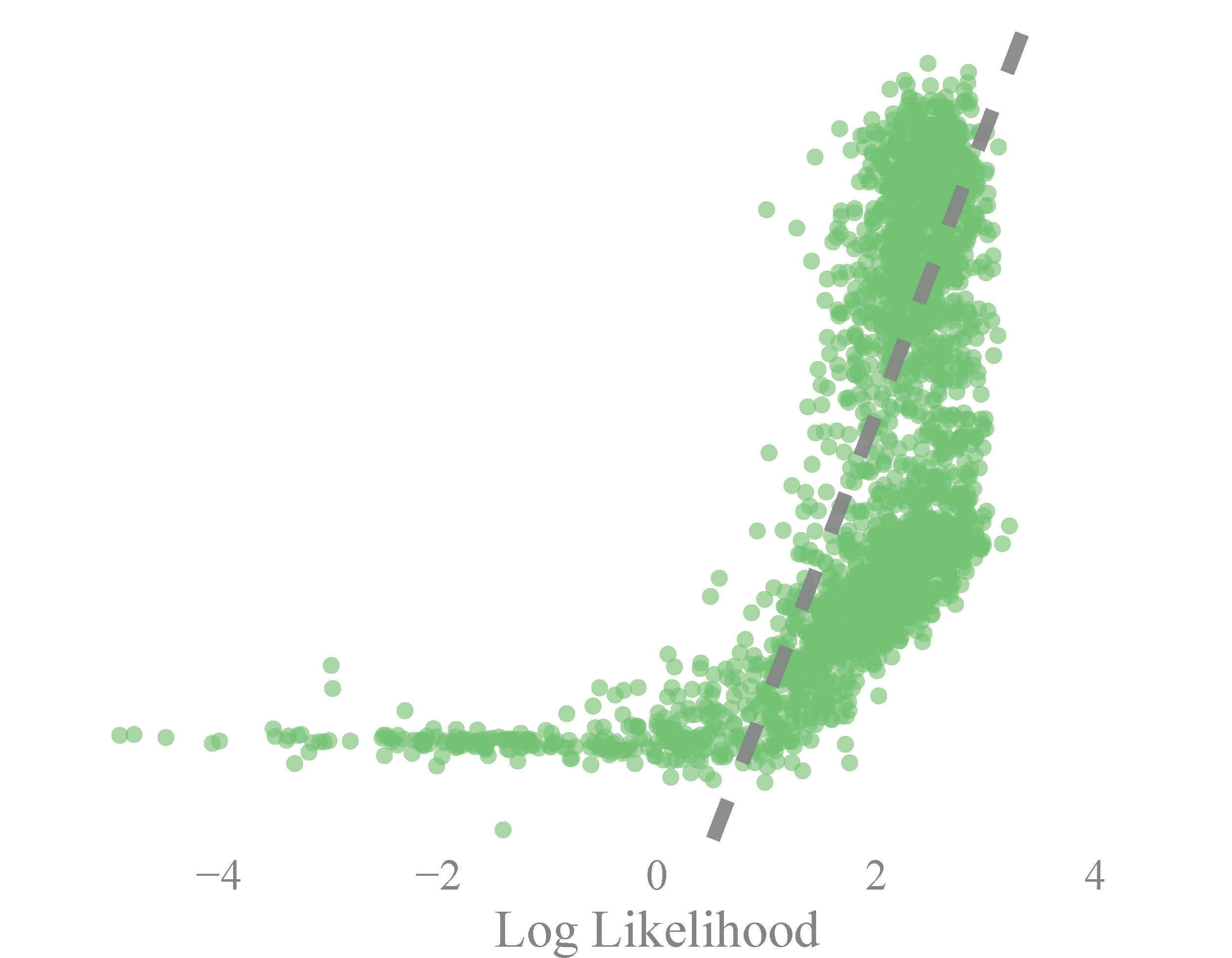}}%
        
      }
\end{figure}

To test this, we re-ran experiments from \sect{sec:nllvr} with the LL being calculated on trajectories rather than random batches. 
\fig{fig:traj-loss} shows an improved trend (less variance in the relationship in the form of a tighter grouping, and increased $\rho$) for cartpole likelihood versus reward in the new data (\fig{fig:cpleft}) over the original experiments (\fig{fig:cpright}).
For halfcheetah, the trajectories are substantially longer ($1000$ timesteps) than the batch size ($64$), so we verify that increasing the batch size of validation is not the only effect in improving the trend of likelihood versus reward.
\fig{fig:hc-sub-batch} shows a tighter relationship between likelihood and reward than the exploration using default PETS values in \fig{fig:hc-sub}. 
Finally, by validating on trajectories versus large batches, the trend of likelihood versus reward is again improved in \fig{fig:hc-sub-traj}.

\section{Using simple reward as re-weight}

\begin{wrapfigure}{r}{.29\textwidth}
    \centering
    \vspace{-10pt}
    \includegraphics[width=\linewidth]{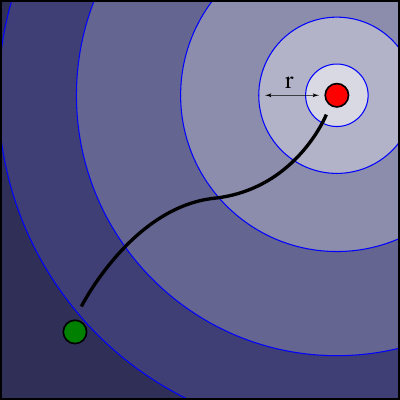}
    \caption{re-weight the loss of the dynamics model w.r.t. the reward.}
    \label{fig:weightrew}
\end{wrapfigure}
An alternative to re-weighting w.r.t. the optimal trajectory could be re-weighting w.r.t. the reward of each state.
The compelling advantage of this would be the easy availability of the reward without access to additional information (e.g., the optimal trajectory).
However, the simple reward does not topologically has the desired shape compared to the optimal trajectory.
In fact, for many rewards (e.g., distance to the target) the isocurves of reward are orthogonal to the optimal trajectory.
This means that the resulting re-weighting would concentrate the dynamics to model accurately the part of the state-action space closer to the target, but it would ignore the dynamics that lead us to the reward in the first space (e.g., along the optimal trajectory).
Intuitively, this is undesirable, as it might decrease performance in the initial stages of the trajectory.
More research will be necessary to fully study alternatives forms of re-weighting.

\section{Ways model mismatch can harm the performance of a controller}
Model mismatch between fitting the likelihood and optimizing
the task's reward manifests itself in many ways.
Here we highlight two of them and in~\sect{sec:Discussion} 
we discuss how related work connects in with these issues.

\paragraph{Long-horizon roll-outs of the model may be unstable and inaccurate.}
Time-series or dynamics models that are unrolled for long
periods of time easily diverge from the true prediction
and can easily step into predicting future states that are not
on the manifold of reasonable trajectories.
Taking these faulty dynamics models and using them as a smaller
part of a controller that optimizes some cost function under
a poor approximation to the dynamics.
Issues can especially manifest if, \eg, the approximate
dynamics do not properly capture stationarity properties
necessary for the optimality of the true physical system
being modeled.

\paragraph{Non-convex and non-smooth models may make the control optimization problem challenging}
The approximate dynamics might have bad properties that make the
control optimization problem much more difficult than on the true system,
even when the true optimal action sequence is optimal under the
approximate model.
This is especially true when using neural network as they introduce
non-linearities and non-smoothness that make many classical control
approaches difficult.

\paragraph{Sampling models with similar LLs, different rewards}
To better understand the objective mismatch, we also compared how a difference of model loss can impact a control policy.
We sampled models with similar LL's and extremely different rewards from \fig{fig:nllVr}d-e and visualized the chosen optimal action sequences along an expert trajectory.
The control policies and dynamics models appear to be converging to different regions of state spaces.
In these visualizations, there is not a emphatic reason why the models achieved different reward, so further study is needed to quantify the impact of model differences. 
The interpretability of the difference between models and controllers will be important to solving the objective-mismatch issue.

\begin{wraptable}{r}{.5\textwidth}
\begin{center}
\vspace{-30pt}
 \begin{tabular}{ l c c} 
 Parameter & Cartpole & Half-Cheetah  \\ 
  \hline \\[0.5ex] 
   \multicolumn{3}{c}{Experiment Parameters}  \\ [0.5ex] 
 \hline 
  Trial Time-steps & 200 & 1000 \\   [1.0ex] 
  
   \multicolumn{3}{c}{Random Sampling Parameters}  \\ [0.5ex] 
 \hline 
  Horizon & 25 & 30 \\ 
 \hline
  Trajectories & 2000 & 2500  \\   [1.0ex] 

   \multicolumn{3}{c}{CEM Parameters}  \\ [0.5ex] 
 \hline 
  Horizon & 25 & 30 \\ 
 \hline
  Trajectories & 400 & 500  \\
 \hline
  Elites & 40 & 50  \\
  \hline
  CEM Iterations & 5 & 5  \\[1.0ex] 
  
 \multicolumn{3}{c}{Network Parameters}  \\ [0.5ex] 
 \hline 
  Width & 500 & 200 \\ 
 \hline
  Depth & 2 & 3  \\ 
  \hline
  E & 5 & 5  \\ [1.0ex] 

 \multicolumn{3}{c}{Training Parameters}  \\ [0.5ex] 
 \hline
 Training Type & Full & Incremental  \\
 \hline
  Full / Initial Epochs  & 100 & 20  \\
 \hline
  Incremental Epochs  & - - & 10  \\
 \hline
 Optimizer & Adam & Adam  \\
 \hline
 Batch Size & 16 & 64  \\ 
 \hline
 Learning Rate & 1E-4 & 1E-4  \\ 
  \hline
 Test Train Split & 0.9 & 0.9 

\end{tabular}
\caption{PETS Hyper-parameters}
\label{tab:paramspets}
\end{center}
\begin{center}
\begin{tabular}{ l c } 
 \hline 
 Type & Number of points \\[0.25ex] 
   \multicolumn{2}{c}{Cartpole Datasets}  \\ [0.5ex] 

  \hline 
  Grid &  16807 \\   
  \hline 
  On-policy &  3780 \\   
  \hline 
  Expert &  2400 \\   [1.0ex] 
  
   \multicolumn{2}{c}{Half Cheetah Datasets}  \\ [0.5ex] 
 \hline 
  Sampled &  200000 \\   
  \hline 
  On-policy &  90900 \\   
  \hline 
  Expert &   3000 
\end{tabular}
\caption{Experimental Dataset Sizes}
\label{tab:dataset}
\end{center}
\vspace{-40pt}
\end{wraptable}

\section{Hyper-paramters and Simulation Environment}
\tab{tab:paramspets} includes the PETS parameters used for our cartpole and half-cheetah experiments. Both of these experiments were run with Mujoco version $1.50.1$ (which we found to be significant in replicating various papers across the field of Deep RL).

\paragraph{Experimental datasets} We include in \tab{tab:dataset} the sizes of each dataset used in the experimental section of this paper. The expert datasets employed are generated by a combination of a) running PETS with a true, environment-based dynamics model for prediction or soft actor-critic at convergence. The on-policy data is taken from the end of a trial that solved the given task (rather then sampling from all on-policy data). The grid dataset for cartpole is generated by slicing the state and action spaces evenly. Due to the high dimensionality of half cheetah, uniform slicing does not work, so the dataset is generated by uniformly sampling within the state and action spaces.

\section{Additional Related Works}
\paragraph{Add inductive biases to the controller}
Prior knowledge can be added to
the controller in the form of hyper-parameters such as
the horizon length, or by penalizing unreasonable
control sequences by using, \eg, a slew rate penalty.
These heuristics can significantly improve the performance
if done correctly but can be difficult to tune.
\cite{jiang2015dependence} use complexity theory
to justify using a \emph{short} planning
horizon with an approximate model to reduce the
the class of induced policies.

\end{document}